\documentclass[runningheads]{llncs}
\usepackage{graphicx}
\usepackage{comment}
\usepackage{amsmath,amssymb} 
\usepackage{color}
\newcommand{\ve}[1]{\mathbf{#1}} 
\newcommand{\ma}[1]{\mathrm{#1}} 
\newcommand{\myparagraph}[1]{\vspace{0.1em}\noindent\textbf{#1}}
\usepackage{arydshln}
\usepackage{subfigure}
\usepackage{pifont}
\usepackage{multirow}
\usepackage{xspace}
\newcommand{\ie}{\textit{i}.\textit{e}.}
\newcommand{\eg}{\textit{e}.\textit{g}.}
\usepackage[colorlinks,hyperindex,breaklinks]{hyperref}
\begin{document}
\pagestyle{headings}
\mainmatter
\def\ECCVSubNumber{}  
\title{Feature Pyramid Transformer} 
\titlerunning{Feature Pyramid Transformer}
\authorrunning{D. Zhang, H. Zhang, \emph{et al}.}
\author{\small{Dong Zhang$^1$\quad
Hanwang Zhang$^2$\quad
Jinhui Tang$^{1}$\thanks{Corresponding author.}\quad
Meng Wang$^3$\quad\\
Xiansheng Hua$^4$\quad
Qianru Sun$^5$
\\
\tt\scriptsize{\{dongzhang, jinhuitang\}@njust.edu.cn \quad \quad hanwangzhang@ntu.edu.sg \quad eric.mengwang@gmail.com \quad xiansheng.hxs@alibaba-inc.com \quad qianrusun@smu.edu.sg}
}}
\institute{
\scriptsize{$^1$School of Computer Science and Engineering, Nanjing University of Science and Technology \\
$^2$Nanyang Technological University \quad
$^3$Hefei University of Technology \\
$^4$Damo Academy, Alibaba Group \quad
$^5$Singapore Management University}}
\maketitle
\begin{abstract}
Feature interactions across space and scales underpin modern visual recognition systems because they introduce beneficial visual contexts. Conventionally, spatial contexts are passively hidden in the CNN's increasing receptive fields or actively encoded by non-local convolution. Yet, the non-local spatial interactions are not across scales, and thus they fail to capture the non-local contexts of objects (or parts) residing in different scales. To this end, we propose a fully active feature interaction across both space and scales, called Feature Pyramid Transformer (FPT). It transforms any feature pyramid into another feature pyramid of the same size but with richer contexts, by using three specially designed transformers in self-level, top-down, and bottom-up interaction fashion. FPT serves as a generic visual backbone with fair computational overhead. We conduct extensive experiments in both instance-level (\ie, object detection and instance segmentation) and pixel-level segmentation tasks, using various backbones and head networks, and observe consistent improvement over all the baselines and the state-of-the-art methods\footnote{Code is open-sourced at https://github.com/ZHANGDONG-NJUST}.
\end{abstract}
\section{Introduction}
\begin{figure}[t!]
\centering
\includegraphics[width=.95 \textwidth]{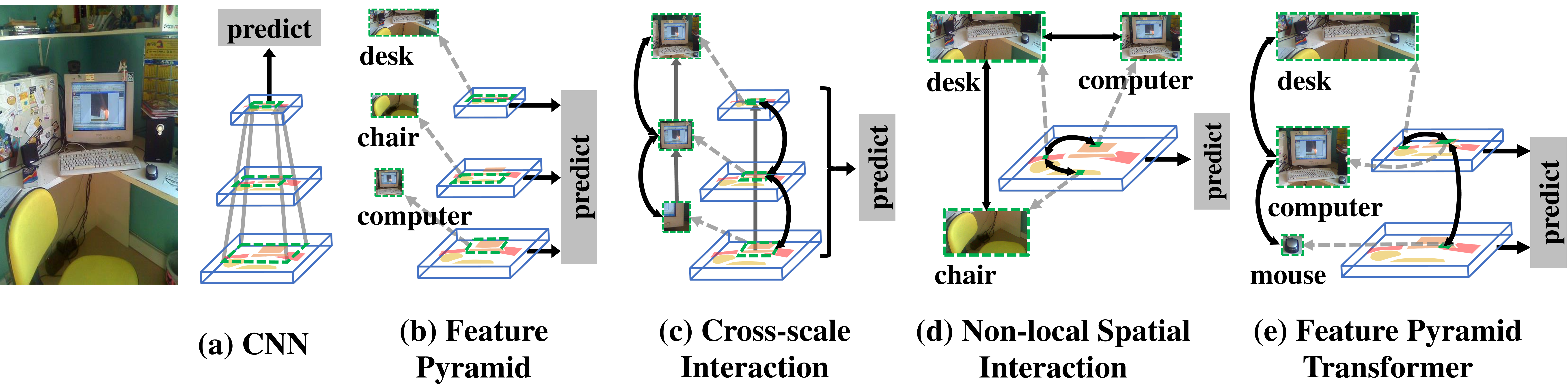}
\caption{The evolution of feature interaction across space and scale in feature pyramid for visual context. Transparent cubes: feature maps. Shaded \texttt{predict}: task-specific head networks. The proposed Feature Pyramid Transformer is inspired by the evolution.}
\label{fig:1}
\end{figure}

Modern visual recognition systems stand in context. Thanks to the hierarchical structure of Convolutional Neural Network (CNN), as illustrated in Fig.~\ref{fig:1}~(a), contexts are encoded in the gradually larger receptive fields (the green dashed rectangles) by pooling~\cite{he2015spatial,lazebnik2006beyond}, stride~\cite{springenberg2014striving} or dilated convolution~\cite{yu2015multi}. Therefore, the prediction from the last feature map is essentially based on the rich contexts --- even though there is only one ``feature pixel'' for a small object, \eg, \texttt{mouse}, its recognition will be still possible, due to the perception of larger contexts, \eg, \texttt{table} and \texttt{computer}~\cite{girshick2015fast,ren2015faster}.

Scale also matters --- the \texttt{mouse} recognition deserves more feature pixels, not only the ones from the last feature map, which easily overlooks small objects. A conventional solution is to pile an \emph{image pyramid} for the same image~\cite{adelson1984pyramid}, where the higher/lower levels are images of lower/higher resolutions. Thus, objects of different scales are recognized in their corresponding levels, \eg, \texttt{mouse} in lower levels (high resolution) and \texttt{table} in higher levels (low resolution). However, the image pyramid multiplies the time-consuming CNN forward pass as each image requires a CNN for recognition. Fortunately, CNN offers an in-network \emph{feature pyramid}~\cite{zeiler2014visualizing}, \ie, lower/higher-level feature maps represent higher/lower-resolution visual content without computational overhead~\cite{liu2016ssd,redmon2016you}. As shown in Fig.~\ref{fig:1}~(b), we can recognize objects of different scales by using feature maps of different levels, \ie, small objects (\texttt{computer}) are recognized in lower-levels and large objects (\texttt{chair} and \texttt{desk}) are recognized in higher-levels~\cite{hu2018relation,lin2017feature,liu2018path}.

Sometimes the recognition --- especially for \emph{pixel-level} labeling such as semantic segmentation --- requires to combine the contexts from multiple scales~\cite{chen2017rethinking,zhao2017pyramid}. For example in Fig.~\ref{fig:1}~(c), to label pixels in the frame area of the \texttt{monitor}, perhaps the local context of the object itself from lower levels is enough; however, for the pixels in the screen area, we need to exploit both of the local context and the global context from higher levels, because the local appearance of \texttt{monitor} screen is close to \texttt{TV} screen, and we should use scene context such as \texttt{keyboard} and \texttt{mouse} to distinguish between the two types.

The spirit of the above non-local context is recently modeled in a more explicit and active fashion --- as opposed to the above passive feature map pile --- by using the non-local convolution~\cite{wang2018non} and self-attention~\cite{vaswani2017attention,carion2020end}. Such spatial feature interaction is expected to capture the reciprocal co-occurring patterns of multiple objects~\cite{zhang2019co,hu2018relation}. As shown in Fig.~\ref{fig:1}~(d), it is more likely that there is a \texttt{computer} on the \texttt{desk} rather than on \texttt{road}, thus, the recognition of either is helpful to the other.

The tale of context and scale should continue, and it is our key motivation. In particular, we are inspired by the omission of the cross-scale interactions (Fig.~\ref{fig:1}~(c)) in the non-local spatial interactions (Fig.~\ref{fig:1}~(d)). Moreover, we believe that the non-local interaction \textit{per se} should happen in the corresponding scales of the interacted objects (or parts), but not just in one uniform scale as in existing methods~\cite{vaswani2017attention,wang2018non,zhang2019co}. Fig.~\ref{fig:1}~(e) illustrates the expected non-local interactions across scales: the low-level \texttt{mouse} is interacting with the high-level \texttt{computer}, which is interacting with \texttt{desk} at the same scale.
\begin{figure}[t]
\centering
\includegraphics[width=.95 \textwidth]{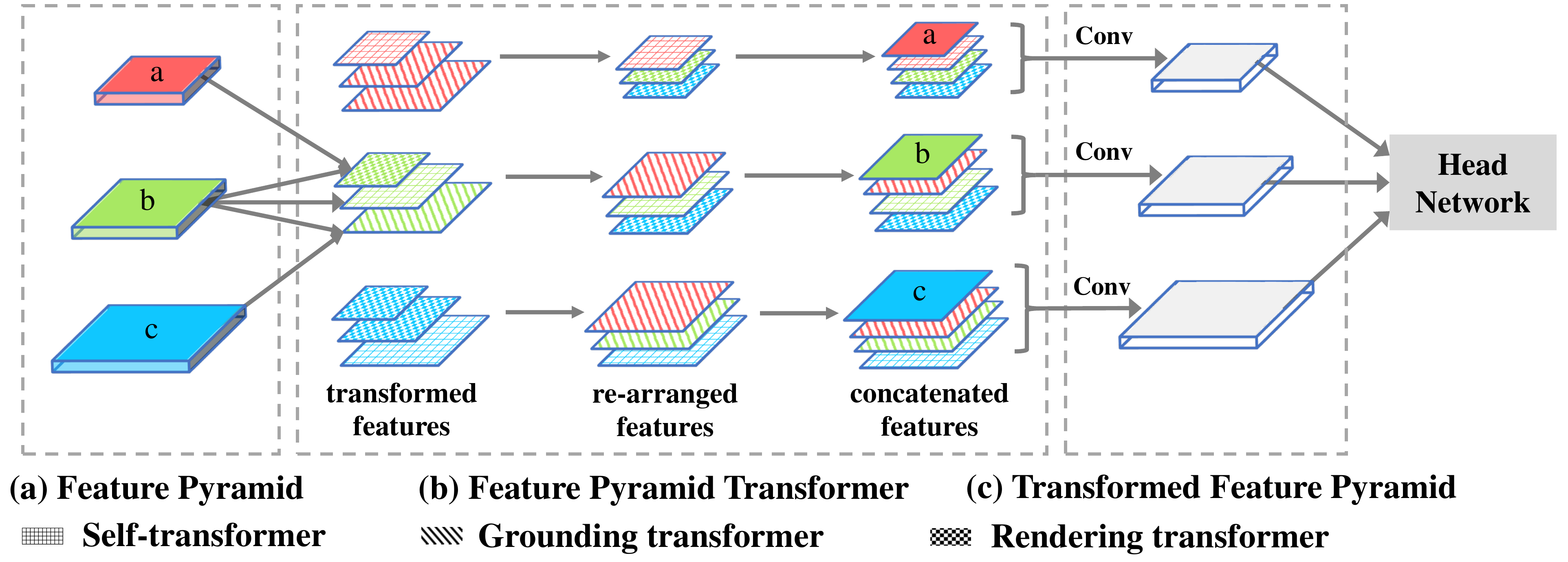}
\caption{Overall structure of our proposed FPT network. Different texture patterns indicate different feature transformers, and different color represents feature maps with different scales. ``Conv'' denotes a $3\times3$ convolution with the output dimension of $256$.
Without loss of generality, the top/bottom layer feature maps has no rendering/grounding transformer.}
\label{fig:2}
\end{figure}

To this end, we propose a novel feature pyramid network called \textbf{Feature Pyramid Transformer} (FPT) for visual recognition, such as instance-level (\ie, object detection and instance segmentation) and pixel-level segmentation tasks. In a nutshell, as illustrated in Fig.~\ref{fig:2}, the input of FPT is a feature pyramid, and the output is a transformed one, where each level is a \emph{richer} feature map that encodes the non-local interactions across space and scales. Then, the feature pyramid can be attached to any task-specific head network. As its name implies, FPT's interaction adopts the transformer-style~\cite{vaswani2017attention,carion2020end}. It has the neat \emph{query}, \emph{key} and \emph{value} operation (cf. Section~\ref{sec:3.1}) that is shown effective in selecting informative long-range interaction, which tailors our goal: non-local interaction at proper scales. In addition, the computation overhead (cf. Section~\ref{sec:4.1}) can be alleviated by using TPUs like any other transformer models~\cite{jouppi2017datacenter}.

Our technical contributions, as illustrated in the FPT breakdown in Fig.~\ref{fig:2}, are the designs of three transformers: 1) \textbf{Self-Transformer} (ST). It is based on the classic non-local interaction within the same level feature map~\cite{wang2018non}, and the output has the same scale as its input. 2) \textbf{Grounding Transformer} (GT). It is in a top-down fashion, and the output has the same scale as the lower-level feature map. Intuitively, we ground the ``concept'' of the higher-level feature maps to the ``pixels'' of the lower-level ones. In particular, as it is unnecessary to use the global information to segment objects, and the context within a local region is empirically more informative, we also design a \emph{locality-constrained} GT for both efficiency and accuracy of semantic segmentation. 3) \textbf{Rendering Transformer} (RT). It is in a bottom-up fashion, and the output has the same scale as the higher-level feature map. Intuitively, we render the higher-level ``concept'' with the visual attributes of the lower-level ``pixels''. Note that this is a \emph{local} interaction as it is meaningless to render an ``object'' with the ``pixels'' of another distant one. The transformed feature maps of each level (the red, blue and green) are re-arranged to its corresponding map size and then concatenated with the original map, before feeding into the conv-layer that resize them to the original ``thickness''.

Extensive experiments show that FPT can greatly improve conventional detection/segmentation pipelines by the following absolute gains: 1) 8.5\% box-AP for object detection and 6.0\% mask-AP for instance segmentation over baseline on the MS-COCO~\cite{lin2014microsoft} \emph{test-dev}; 2) for semantic segmentation, 1.6\% and 1.2\% mIoU on Cityscapes~\cite{cordts2016cityscapes} and PASCAL VOC 2012~\cite{everingham2015pascal} test sets, respectively; 1.7\% and 2.0\% mIoU on ADE20K~\cite{zhou2017scene} and LIP~\cite{gong2017look} validation sets, respectively.
\section{Related Work}
FPT is generic to apply in a wide range of computer vision tasks. This paper focuses on two instance-level tasks: object detection, instance segmentation, and one pixel-level task: semantic segmentation. Object detection aims to predict a bounding box for each object and then assigns the bounding box a class label~\cite{ren2015faster}, while instance segmentation is additionally required to predict a pixel-level mask of the object~\cite{he2017mask}. Semantic segmentation aims to predict a class label to each pixel of the image~\cite{long2015fully}.

\myparagraph{Feature pyramid.}
The in-network feature pyramid (\ie, the Bottom-up Feature Pyramid (BFP)~\cite{lin2017feature}) is one of the most commonly used methods, and has been shown useful for boosting object detection~\cite{liu2016ssd}, instance segmentation~\cite{liu2018path} and semantic segmentation~\cite{zhang2018exfuse}.
Another popular method of constructing feature pyramid uses feature maps of the scale while processing the maps through pyramidal pooling or dilated/atrous convolutions. For example, atrous spatial pyramid pooling~\cite{chen2017rethinking} and pyramid pooling module~\cite{he2015spatial,zhao2017pyramid} leverages output feature maps of the last convolution layer in the CNN backbone to build the four-level feature pyramid, in which different levels have the same resolution but different information granularities. Our approach is based on the existing BFP (for the instance-level) and unscathed feature pyramid~\cite{peng2017large} (for the pixel-level). Our contribution is the novel feature interaction approach.

\myparagraph{Feature interaction.}
An intuitive approach to the cross-scale feature interaction is gradually summing the multi-scale feature maps, such as Feature Pyramid Network (FPN)~\cite{lin2017feature} and Path Aggregation Network (PANet)~\cite{liu2018path}. In particular, both FPN and PANet are based on BFP, where FPN adds a top-down path to propagate semantic information into low-level feature maps, and PANet adds a bottom-up path augmentation on the basis of FPN. Another approach is to concatenate multi-scale feature maps along the channel dimension. The specific examples for semantic segmentation are DeepLab~\cite{chen2017deeplab} and pyramid scene parsing network~\cite{zhao2017pyramid}. Besides, a more recent work proposed the ZigZagNet~\cite{lin2019zigzagnet} which exploits the addition and convolution to enhance the cross-scale feature interaction. Particularly, for the within-scale feature interaction, some recent works exploited non-local operation~\cite{wang2018non} and self-attention~\cite{vaswani2017attention} to capture the co-occurrent object features in the same scene. Their models were evaluated in a wide range of visual tasks~\cite{hu2018relation,yuan2018ocnet,zhang2019co,zhu2019asymmetric}. However, we argue that the non-local interaction performed in just one uniform scale feature map is not enough to represent the contexts. In this work, we aim to conduct the non-local interaction \textit{per se} in the corresponding scales of the interacted objects (or parts).
\section{Feature Pyramid Transformer}
Given an input image, we can formally extract a feature pyramid, where the fine-/coarse-grained feature maps are in low/high levels, respectively. Without loss of generality, we express a low-level fine-grained feature map as $\ve{X}^f$ and a high-level coarse-grained feature map as $\ve{X}^c$. \textbf{Feature Pyramid Transformer} (FPT) enables features to interact across space and scales. It specifically includes three transformers: self-transformer (cf. Section~\ref{sec:3.2}), grounding transformer (cf. Section~\ref{sec:3.3}) and rendering transformer (cf. Section~\ref{sec:3.4}). The transformed feature pyramid is in the same size but with richer contexts than the original.
\subsection{Non-Local Interaction Revisited}\label{sec:3.1}
A typical non-local interaction~\cite{wang2018non} operates on $\emph{queries} (\ma{Q})$, $\emph{keys} (\ma{K})$ and $\emph{values} (\ma{V})$ within a single feature map $\ve{X}$, and the output is the transformed version $\tilde{\ve{X}}$ with the same scale as $\ve{X}$. This non-local interaction is formulated as:
\begin{equation}
\small
\begin{split}
\textbf{Input:} \quad & \ve{q}_i, \ve{k}_j, \ve{v}_j \\
\textbf{Similarity:}\quad & s_{i,j} = F_{sim}(\ve{q}_i, \ve{k}_j) \\
\textbf{Weight:}\quad & w_{i,j} = F_{nom}(s_{i,j}) \\
\textbf{Output:}\quad &
\tilde{\ve{X}}_i = F_{mul}(w_{i,j}, \ve{v}_j),
\label{eq:1}
\end{split}
\end{equation}
where $\ve{q}_i=f_q(\ve{X}_i)\in\ma{Q}$ is the $i^{th}$ \emph{query}; $\ve{k}_j=f_k(\ve{X}_j)\in\ma{K}$ and $\ve{v}_j=f_v(\ve{X}_j)\in\ma{V}$ are the $j^{th}$ \emph{key}/\emph{value} pair; $f_q(\cdot)$, $f_k(\cdot )$ and $f_v(\cdot )$ denote the \emph{query}, \emph{key} and \emph{value} transformer functions~\cite{carion2020end,vaswani2017attention}, respectively. $\ve{X}_i$ and $\ve{X}_j$ are the $i^{th}$ and $j^{th}$ feature positions in $\ve{X}$, respectively. $F_{sim}$ is the similarity function (default as \emph{dot product}); $F_{nom}$ is the normalizing function (default as \emph{softmax}); $F_{mul}$ is the weight aggregation function (default as \emph{matrix multiplication}); and $\tilde{\ve{X}}_i$ is the $i^{th}$ feature position in the transformed feature map $\tilde{\ve{X}}$.
\subsection{Self-Transformer}\label{sec:3.2}
\textbf{Self-Transformer} (ST) aims to capture the co-occurring object features on one feature map. As illustrated in Fig.~\ref{fig:4} (a), ST is a modified non-local interaction~\cite{wang2018non} and the output feature map $\hat{\ve{X}}$ has the same scale as its input $\ve{X}$.
A main difference with~\cite{vaswani2017attention,wang2018non} is that we deploy the
Mixture of Softmaxes (MoS)~\cite{yang2017breaking} as the normalizing function $F_{mos}$,
which turns out to be more effective than the standard $\emph{Softmax}$~\cite{zhang2019co} on images. Specifically, we first divide $\ve{q}_i$ and $\ve{k}_j$ into $\mathcal{N}$ parts. Then, we calculate a similarity score $s^n_{i,j}$ for every pair, \ie, $\ve{q}_{i,n}$, $\ve{k}_{j,n}$, using $F_{sim}$.
The MoS-based normalizing function $F_{mos}$ is as follows:
\begin{equation}
\small
\begin{aligned}
F_{mos}(s^n_{i,j}) = \sum_{n=1}^{\mathcal{N}}\pi_{n}\frac{\mathrm{exp}(s^n_{i,j})}{\sum_j \mathrm{exp}(s^n_{i,j})},
\label{eq:2}
\end{aligned}
\end{equation}
where $s^n_{i,j}$ is the similarity score of the $n^{th}$ part.
$\pi_{n}$ is the $n^{th}$ aggregating weight
that is equal to  $\emph{Softmax}\left(\ve{w}^T_n\bar{\ve{k}}\right)$,
where $\ve{w}_n$ is a learnable linear vector for normalization and $\bar{\ve{k}}$ is the arithmetic mean of all positions of $\ve{k}_j$.
Based on $F_{mos}$,
we then can reformulate Eq.~\ref{eq:1} to elaborate our proposed ST as follows:
\begin{equation}
\small
\begin{split}
\textbf{Input:} \quad & \ve{q}_i, \ve{k}_j, \ve{v}_j, \mathcal{N} \\
\textbf{Similarity:}\quad & s^n_{i,j} = F_{sim}(\ve{q}_{i,n}, \ve{k}_{j,n}) \\
\textbf{Weight:}\quad & w_{i,j} = F_{mos}(s^n_{i,j}) \\
\textbf{Output:}\quad & \hat{\ve{X}}_i = F_{mul}(w_{i,j}, \ve{v}_j),
\label{eq:3}
\end{split}
\end{equation}
where $\hat{\ve{X}}_i$ is the $i^{th}$ transformed feature position in $\hat{\ve{X}}$.

\subsection{Grounding Transformer}\label{sec:3.3}
\textbf{Grounding Transformer} (GT) can be categorized as a top-down non-local interaction~\cite{wang2018non}, which grounds the ``concept'' in the higher-level feature maps $\ve{X}^c$ to the ``pixels'' in the lower-level feature maps $\ve{X}^f$. The output $\hat{\ve{X}}^f$ has the same scale as $\ve{X}^f$.
Generally, image features at different scales extract different semantic or contextual information or both~\cite{zeiler2014visualizing,zhang2018exfuse}. Moreover, it has been empirically shown that the negative value of the \emph{euclidean distance} $F_{eud}$ is more effective in computing the similarity than \emph{dot product} when the semantic information of two feature maps is different~\cite{zhang2018learning}. So we prefer to use $F_{eud}$ as the similarity function, which is expressed as:
\begin{equation}
\small
\begin{aligned}
F_{eud}(\ve{q}_i, \ve{k}_j) = -||\ve{q}_{i}-\ve{k}_{j}||^2,
\label{eq:4}
\end{aligned}
\end{equation}
where $\ve{q}_i=f_q(\ve{X}^f_i)$ and $\ve{k}_j=f_k(\ve{X}^c_j)$; $\ve{X}^f_i$ is the $i^{th}$ feature position in $\ve{X}^f$, and $\ve{X}^c_j$ is the $j^{th}$ feature position in $\ve{X}^c$. We then replace the similarity function in Eq.~\ref{eq:3} with
$F_{eud}$, and get the formulation of
the proposed
GT as follows:
\begin{equation}
\small
\begin{aligned}
\textbf{Input:} \quad & \ve{q}_i, \ve{k}_j, \ve{v}_j, \mathcal{N} \\
\textbf{Similarity:}\quad & s^n_{i,j} = F_{eud}(\ve{q}_{i,n}, \ve{k}_{j,n})\\
\textbf{Weight:}\quad & w_{i,j} = F_{mos}(s^n_{i,j}) \\
\textbf{Output:}\quad & \hat{\ve{X}}^f_i = F_{mul}(w_{i,j}, \ve{v}_j),
\label{eq:5}
\end{aligned}
\end{equation}
where $\ve{v}_j=f_v(\ve{X}^c_j)$; $\hat{\ve{X}}^f_i$ is the $i^{th}$ transformed feature position in $\hat{\ve{X}}^f$.
Based on Eq.~\ref{eq:5}, each
pair of $\ve{q}_i$ and $\ve{k}_j$
with a closer distance
will be
given a larger weight
as in~\cite{vaswani2017attention,wang2018non}.
Compared to the results of \emph{dot product}, using $F_{eud}$ brings
clear improvements in the top-down interactions\footnote{More details are given in \emph{Section A} of the supplementary.}.

\begin{figure}[t]
\centering
\includegraphics[width=.95 \textwidth]{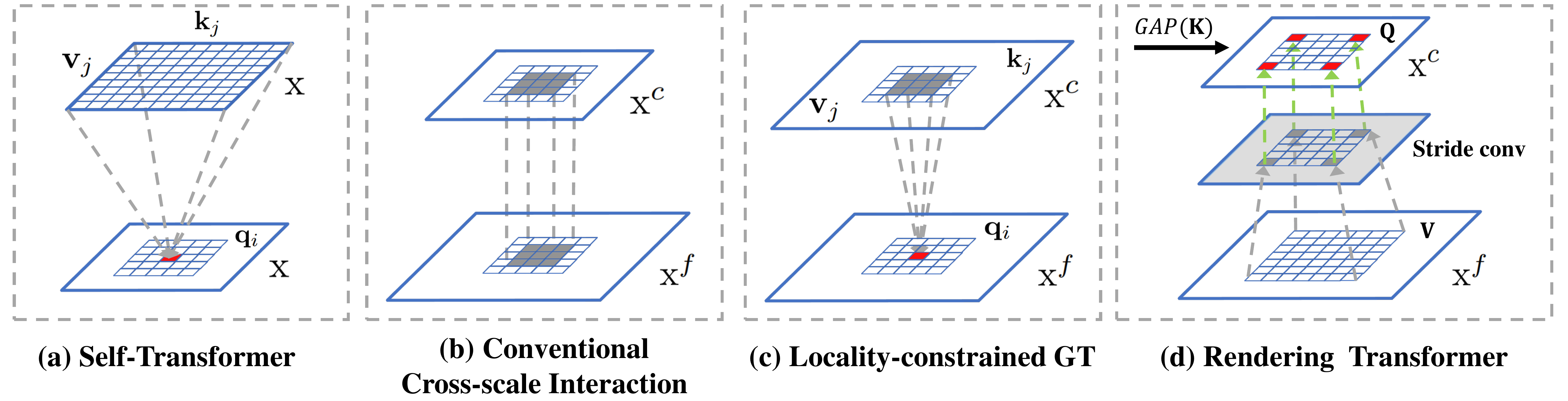}
\caption{Self-Transformer(ST), Conventional Cross-Scale Interaction in existing methods, Locality-constrained Grounding Transformer (GT), and Rendering Transformer. The red grid in low-level is a \emph{query} position; grids in high-level are the \emph{key} and the \emph{value} positions (within a local square area in (b)); $\textbf{Q}$ are the high-level feature maps, $\textbf{K}$ and $\textbf{V}$ are the low-level feature maps. Grey square is the down-sampled $\textbf{V}$.}
\label{fig:4}
\end{figure}

In feature pyramid, high-/low-level feature maps contain much global/local image information. However, for semantic segmentation by cross-scale feature interactions, it is unnecessary to use global information to segment two objects in an image. The context within a local region around the \emph{query} position is empirically more informative. That is why the conventional cross-scale interaction (\eg, summation and concatenation) is effective in existing segmentation methods~\cite{chen2017deeplab,zhao2017pyramid}. As shown in Fig.~\ref{fig:4} (b), they are essentially the implicit local style. However, our default GT is the global interaction.

\myparagraph{Locality-constrained Grounding Transformer.}
We therefore introduce a \emph{locality-constrained} version of GT called Locality-constrained GT (LGT) for semantic segmentation, which is an explicit local feature interaction. As illustrated in Fig.~\ref{fig:4} (c), each $\ve{q}_i$ (\ie, the red grid on the low-level feature map) interacts with a portion of $\ve{k}_j$ and $\ve{v}_j$ (\ie, the blue grids on the high-level feature map) within the local square area where the center coordinate is the same with $\ve{q}_i$ and the side length is $\emph{square\_size}$. Particularly, for positions of $\ve{k}_j$ and $\ve{v}_j$ that exceed the index, we use $0$ value instead.

\subsection{Rendering Transformer}\label{sec:3.4}
\textbf{Rendering Transformer} (RT) works in a bottom-up fashion, aiming to render the high-level ``concept'' by incorporating the visual attributes in the low-level ``pixels''.
As illustrated in Fig.~\ref{fig:4}~(d), RT is a \emph{local} interaction where the \emph{local} is due to the fact that it is meaningless to render an ``object'' with the features or attributes from another distant one.

In our implementation, RT is not performed by pixel but the entire feature maps.
Specifically, the high-level feature map is defined as $\textbf{Q}$; the low-level feature map is defined as $\textbf{K}$ and $\textbf{V}$. To highlight the rendering target, the interaction between $\textbf{Q}$ and $\textbf{K}$ is conducted in a channel-wise attention manner~\cite{chen2017sca}. $\textbf{K}$ first computes a weight $\ve{w}$ for $\textbf{Q}$ through Global Average Pooling (GAP)~\cite{lin2013network}. Then, the weighted $\textbf{Q}$ (\ie, $\textbf{Q}_{att}$) goes through a $3 \times 3$ convolution for refinement~\cite{yu2018learning}. $\textbf{V}$ goes through a $3 \times 3$ convolution with stride to reduce the feature scale (the gray square in Fig.~\ref{fig:4}~(d)). Finally, the refined $\textbf{Q}_{att}$ and the down-sampled $\textbf{V}$ (\ie, $\textbf{V}_{dow}$) are summed-up, and processed by another $3 \times 3$ convolution for refinement. The proposed RT can be formulated as follows:
\begin{equation}
\small
\begin{aligned}
\textbf{Input:} \quad & \textbf{Q}, \textbf{K}, \textbf{V} \\
\textbf{Weight:}\quad & \ve{w} = \emph{GAP}(\textbf{K}) \\
\textbf{Weight Query:}\quad & \textbf{Q}_{att} = F_{att}(\textbf{Q}, \ve{w}) \\
\textbf{Down-sampled Value:} \quad & \textbf{V}_{dow} =  F_{sconv} (\textbf{V}) \\
\textbf{Output:}\quad & \hat{\textbf{X}}^c = F_{add} (F_{conv}(\textbf{Q}_{att}),\textbf{V}_{dow}),
\label{eq:6}
\end{aligned}
\end{equation}
where $F_{att}(\cdot)$ is an \emph{outer product} function; $F_{sconv}(\cdot)$ is
a $3 \times 3$ stride convolution, in particular, where $stride=1$ if the scales of $\textbf{Q}$ and $\textbf{V}$ are equal; $F_{conv}(\cdot)$ is a $3 \times 3$ convolution for refinement; $F_{add}(\cdot)$ is the feature map summation function with a $3 \times 3$ convolution; and $\hat{\textbf{X}}^c$ denotes the output feature map of RT.

\subsection{Overall Architecture}\label{sec:3.5}
We build specific FPT networks for tackling object detection~\cite{hu2018relation,lin2017feature}, inatance segmentation~\cite{he2017mask,liu2018path}, and semantic segmentation~\cite{chen2017rethinking,zhao2017pyramid}. Each FPT network is composed of four components: a backbone for feature extraction; a feature pyramid construction module; our proposed FPT for feature transformer; and a task-specific head network.
In the following, we detail the proposed architectures.

\myparagraph{FPT for object detection and instance segmentation.}
We follow~\cite{lin2017feature,liu2018path} to deploy the ResNet as the backbone, and pre-train it on the ImageNet~\cite{deng2009imagenet}. BFP~\cite{lin2017feature} is used as the pyramid construction module.
Then the proposed FPT is applied to BFP, for which the number of divided parts of $\mathcal{N}$ is set to $2$ for ST and $4$ for GT\footnote{More details are given in \emph{Section B} of the supplementary.}.
Then, the transformed feature maps (by FPT) are concatenated with the original maps along the channel dimension. The concatenated maps go through a $3 \times 3$ convolution to reduce the feature dimension into $256$. On the top of the output feature maps, we apply the head networks for handling specific tasks, \eg, the Faster R-CNN~\cite{ren2015faster} head for object detection and the Mask R-CNN~\cite{he2017mask} head for instance segmentation. To enhance the feature generalization, we apply the DropBlock~\cite{ghiasi2018dropblock} to each output feature map. We set the drop block size as $5$ and the feature keep probability as $0.9$.

\myparagraph{FPT for semantic segmentation.}
We use dilated ResNet-101~\cite{yu2015multi} as the backbone (pre-trained on the ImageNet) following~\cite{chen2017rethinking,zhang2018context}.
We then apply the Unscathed Feature Pyramid (UFP) as the feature pyramid construction module, which basically contains a pyramidal global convolutional network~\cite{peng2017large} with the internal kernel size of $1$, $7$, $15$ and $31$, and each scale with the output dimension of $256$.
Then, the proposed FPT (including LGT) is applied to UFP with the same number of divided parts $\mathcal{N}$ as in the instance-level tasks. In particular, the $\emph{square\_size}$ of LGT is set to 5. On the top of the transformed feature pyramid, we apply the semantic segmentation head network, as in~\cite{chen2017rethinking,zhang2019co}. We also deploy the DropBlock~\cite{ghiasi2018dropblock} on the output feature maps with the drop block size as $3$ and the feature keep probability as $0.9$.
\section{Experiments}
Our experiments were conducted on three interesting and challenging tasks: \ie, instance-level object detection and segmentation, and pixel-level semantic segmentation. In each task, we evaluated our approach with careful ablation studies, extensive comparisons to the state-of-the-arts and representative visualizations.
\subsection{Instance-Level Recognition}\label{sec:4.1}
\myparagraph{Dataset.}
Experiments on object detection and instance segmentation were conducted on MS-COCO 2017~\cite{lin2014microsoft} which has $80$ classes and includes
$115$k, $5$k and $20$k images for training, validation and test, respectively.

\myparagraph{Backbone.}
In the ablation study, ResNet-50~\cite{he2016deep} was used as the backbone.
To compare to state-of-the-arts, we also employed ResNet-101~\cite{he2016deep}, Non-local Network (NL-ResNet-101)~\cite{wang2018non}, Global Context Network (GC-ResNet-101)~\cite{cao2019gcnet} and Attention Augmented Network (AA-ResNet-101)~\cite{IrwanBello2019aanet} as the backbone networks.

\myparagraph{Setting.}
As in~\cite{lin2017feature,liu2018path}, the backbone network was pre-trained on the ImageNet~\cite{deng2009imagenet}, then the whole network was fine-tuned on the training data while freezing the backbone parameters. For fair comparisons, input images were resized into $800$ pixels/$1,000$ pixels for the shorter/longer edge~\cite{lin2019zigzagnet}.

\myparagraph{Training details.}
We adopted SGD training on $8$ GPUs with the Synchronized Batch Norm (SBN)~\cite{zhang2018context}. Each mini-batch involved one image per GPU and $512$ Region of Interest (ROI) per image. The positive-to-negative ratio was set to $1:3$. The weight decay and momentum were set to $0.0001$ and $0.9$, respectively. For object detection, the learning rate was $0.05$ in the first $80$k iterations, and $0.005$ in the remaining $20$k iterations. For instance segmentation, the learning rate was $0.05$ for the first $120$k iterations, and $0.005$ in the remaining $40$k iterations. An end-to-end region proposal network was used to generate proposals, as in~\cite{wang2018non}.

\myparagraph{Comparison methods.}
We compared our FPT to the state-of-the-art cross-scale feature pyramid interactions including FPN~\cite{lin2017feature}, Bottom-up Path Aggregation (BPA) in PANet~\cite{liu2018path}, and Bi-direction Feature Interaction (BFI) in ZigZagNet~\cite{lin2019zigzagnet}.
We also reported the experimental results of using the Augmented Head (AH)~\cite{liu2018path} and Multi-scale Training (MT)~\cite{liu2018path}, where the AH specifically includes the adaptive feature pooling, fully-connected fusion, and heavier head.

\myparagraph{Metrics.}
We evaluated the model performance using the standard Average Precision (AP), AP$_{50}$, AP$_{75}$, AP$_S$, AP$_M$ and AP$_L$.

\myparagraph{Ablation study.}
Our ablation study aims to (1)~evaluate the performance of three individual transformers (in our FPT) and combinations, for which the base pyramid method BFP~\cite{lin2017feature} is the baseline (in Table~\ref{tab1}), and (2)~investigate the effects of SBN~\cite{zhang2018context} and DropBlock~\cite{ghiasi2018dropblock} on our FPT (in Table~\ref{tab:r1}).

\myparagraph{Comparing to the baseline.}
Table~\ref{tab1} show that three transformers bring consistent improvements over the baseline. For example, ST, GT and RT respectively brings $0.4\%$, $3.5\%$ and $3.1\%$ improvements for the bounding box AP in object detection. The improvements are higher as $0.7\%$, $4.0\%$ and $3.2\%$ for the mask AP in instance segmentation. The gain by ST is not as much as the gains by the other two transformers. An intuitive reason is that, compared to self-interaction (\ie, ST), the cross-scale interactions (\ie, GT and RT) capture more diverse and richer inter-object contexts to achieve better object recognition and detection performances, which is consistent with the conclusion of instance-level recognition works~\cite{zhou2018weakly,zhu2019portrait}. The middle blocks in Table~\ref{tab1} show that the combination of transformers improves the performance over individuals in most of cases. In particular, the full combination of ST, GT and RT results the best performance, \ie, $38.0\%$ bounding box AP ($6.4\%$ higher than BFP) on object detection and $36.8\%$ mask AP ($6.9\%$ higher than BFP) on instance segmentation.
\begin{table*}[t]
\scriptsize
\begin{center}
\renewcommand\arraystretch{1.3}
\setlength{\tabcolsep}{0.35pt}{
\begin{tabular}{cccc|c:cc:cc:cc:cc:cc:c|cc}
BFP & ST & GT & RT & \multicolumn{2}{c}{{AP}} & \multicolumn{2}{c}{{AP}$_{50}$} & \multicolumn{2}{c}{{AP}$_{75}$} & \multicolumn{2}{c}{{AP}$_S$} & \multicolumn{2}{c}{{AP}$_M$} & \multicolumn{2}{c|}{{AP}$_L$} & Params & GFLOPs \\ \hline \hline
\checkmark & \ding{55} & \ding{55} & \ding{55} & 31.6 & 29.9 & 54.1 & 50.7 & 35.9 & 34.7 & 16.1 & 14.2 & 32.5 & 31.6 & 48.8 & 48.5 & 34.6 M & 172.3 \\ \hline
\checkmark & \checkmark & \ding{55} & \ding{55} & 32.0 & 30.6 & 54.9 & 51.4 & 36.9 & 35.5 & 16.5 & 15.1 & 34.0 & 32.1 & 49.1 & 49.7 & 55.0 M & 248.2  \\
\checkmark & \ding{55} & \checkmark & \ding{55} & 35.1 & 33.9 & 55.2 & 52.4 & 38.1 & 37.7 & 17.4 & 16.9 & 36.3 & 33.3 & 50.3 & 51.7 & 63.9 M & 265.1 \\
\checkmark & \ding{55} & \ding{55} & \checkmark & 34.7 & 33.1 & 55.5 & 52.0 & 37.5 & 37.7 & 17.0 & 15.3 & 36.6 & {34.9} & 52.0 & 52.1 & 39.8 M & 187.9 \\ \hline
\checkmark & \checkmark & \checkmark & \ding{55} & 35.7 & 34.6 & 55.7 & 54.1 & 38.3 & 37.9 & 18.0 & 17.4 & 36.5 & 34.0 & 52.1 & 50.5 & 82.5 M & 322.9 \\
\checkmark & \checkmark & \ding{55} & \checkmark & 35.9 & 34.4 & 56.8 & 55.1 & 38.8 & 38.0 & 19.1 & 17.9 & 37.0 & 34.8 & 53.1 & 52.2 & 61.2 M & 256.7  \\
\checkmark & \ding{55} & \checkmark & \checkmark & {36.9} & {35.1} & {56.6} & {54.5} & {38.2} & {38.5} & {18.8} & {17.7} & {37.7} & \textbf{35.3} & {54.3} & {53.2} & 69.6 M & 281.6 \\
\checkmark & \checkmark & \checkmark & \checkmark & \textbf{38.0} & \textbf{36.8} & \textbf{57.1} & \textbf{55.9} & \textbf{38.9} & \textbf{38.6} & \textbf{20.5} & \textbf{18.8} & \textbf{38.1} & \textbf{35.3} & \textbf{55.7}& \textbf{54.2} & 88.2 M & 346.2  \\ \hline
\multicolumn{4}{c|}{improvements} & $\uparrow$~\textbf{6.4}&$\uparrow$~\textbf{6.9} & $\uparrow$~\textbf{3.0}&$\uparrow$~\textbf{5.2} & $\uparrow$~\textbf{3.0}&$\uparrow$~\textbf{3.9} & $\uparrow$~\textbf{4.4}&$\uparrow$~\textbf{4.6} & $\uparrow$~\textbf{5.6}&$\uparrow$~\textbf{3.7} & $\uparrow$~\textbf{6.9}&$\uparrow$~\textbf{5.7} & ~~ & ~~ \\
\end{tabular}}
\end{center}
\caption{Ablation study on MS-COCO 2017 val set~\cite{lin2014microsoft}. ``BFP'' is Bottom-up Feature Pyramid~\cite{lin2017feature}; ``ST'' is Self-Transformer; ``GT'' is Grounding Transformer; ``RT'' is Rendering Transformer. Results on the left and right of the dashed are of bounding box detection and instance segmentation.}
\label{tab1}
\end{table*}

\myparagraph{Effects of SBN and DropBlock.}
Table~\ref{tab:r1} shows that both SBN and DropBlock improve the model performance of FPT. Their combination yields $0.8\%$ of improvement for the bounding box AP in object detection, and $0.9\%$ for the mask AP in instance segmentation.
\begin{table*}[t]
\scriptsize
\begin{center}
\renewcommand\arraystretch{1.2}
\setlength{\tabcolsep}{2.5pt}{
\begin{tabular}{ccc|c:cc:cc:cc:cc:cc:c}
FPT & SBN & DropBlock & \multicolumn{2}{c}{AP} & \multicolumn{2}{c}{AP$_{50}$} & \multicolumn{2}{c}{AP$_{75}$} & \multicolumn{2}{c}{AP$_S$} & \multicolumn{2}{c}{AP$_M$} & \multicolumn{2}{c}{AP$_L$}\\ \hline \hline
\checkmark& \ding{55} & \ding{55} & 37.2 & 35.9 & 56.0 & 54.3 & 37.7 & 36.9 & 19.0 & 17.2 & 37.7 & 34.8 & 53.1 & 51.3 \\
\checkmark & \checkmark & \ding{55} & 37.8 & 36.5 & 56.7 & 55.2 & 38.4 & 38.2 & 19.6 & 18.0 & 37.9 & 35.1 & 54.0 & 52.1 \\
\checkmark & \ding{55} & \checkmark & 37.5 & 36.2 & 56.5 & 54.8 & 38.0 & 37.3 & 19.5 & 17.8 & 37.8 & 35.0 & 53.8 & 51.9 \\
\checkmark & \checkmark & \checkmark &\textbf{38.0} & \textbf{36.8} & \textbf{57.1} & \textbf{55.9} & \textbf{38.9} & \textbf{38.6} & \textbf{20.5} & \textbf{18.8} & \textbf{38.1} & \textbf{35.3} & \textbf{55.7}& \textbf{54.2} \\
\end{tabular}}
\end{center}
\caption{
{Ablation study of SBN~\cite{zhang2018context} and DropBlock~\cite{ghiasi2018dropblock} on the MS-COCO 2017 val set~\cite{lin2014microsoft}. Results on the left and right of dashed lines are respectively for bounding box detection and instance segmentation.}}
\label{tab:r1}
\end{table*}

\myparagraph{Model efficiency\footnote{More details are given in the \emph{Section C} of the supplementary.}.} We reported the model Parameters (Params) and GFLOPs with the Mask R-CNN~\cite{he2017mask}. Adding +ST, +GT and +RT to the baseline respectively increase Params by $0.59 \times$, $0.85 \times$ and $0.15 \times$ (with mask AP improvements of 0.7\%, 4.0\%, 3.2\%). Correspondingly, GFLOPs are increased by $0.44 \times$, $0.54 \times$ and $0.09 \times$. Compared to related works~\cite{lin2019zigzagnet,lin2017feature,wang2018non}, these are relatively fair overheads on average. For example, the classical Non-local~\cite{wang2018non} has $0.24 \times$ Params for instance segmentation with only $0.9\%$ mask AP improvements on ResNet-50.
\begin{figure}[t]
\centering
\includegraphics[width=.9 \textwidth]{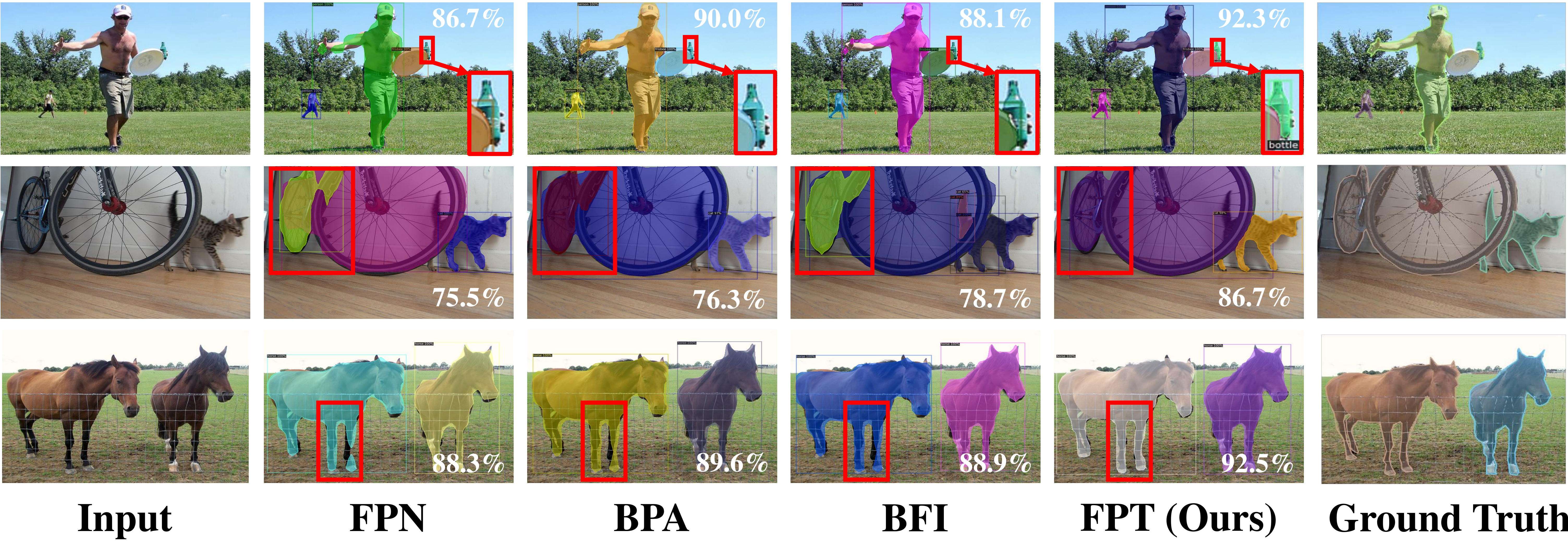}
\caption{Visualization results in instance segmentation. The red rectangle highlights the better predicted area of FPT. Samples are from MS-COCO 2017 validation set~\cite{lin2014microsoft}. The value on each image represents the corresponding segmentation mIoU.}
\label{fig:5}
\end{figure}

\myparagraph{Comparing to the state-of-the-arts.}
The top three blocks in Table~\ref{tab2} show that applying the cross-scale interaction methods, \eg, FPN~\cite{lin2017feature}, BPA~\cite{liu2018path}, BFI~\cite{lin2019zigzagnet} and FPT, results consistent improvements over the baseline BFP~\cite{lin2017feature}. In particular, our FPT achieves the highest gains, \ie, $8.5\%$ AP in object detection and $6.0\%$ mask AP in instance segmentation, with the ResNet-101 backbone~\cite{he2016deep}. Besides, the consistent improvements are also achieved on the stronger NL-, GC- and AA- ResNet-101, and validate that BFP+FPT can generalize well to stronger backbones, which make more senses in \emph{the age of results}\footnote{More results are given in \emph{Section D} of the supplementary.}. At last but not the least, two bottom blocks in Table~\ref{tab2} show that adding efficient training strategies such as AH, MT, and both (denoted as ``[all]'' in Table~\ref{tab2}) to BFP+FPT yields performance boosts. For example, BFP+FPT [all] (with ResNet-101) achieves a higher bounding box AP in object detection and the same mask AP in instance segmentation, compared to the best performance of BFP+FPT (with stronger GC-ResNet-101). Besides, BFP+FPT [all] achieves the average $1.5\%$ AP in object detection and $2.1\%$ mask AP in instance segmentation (over BFP+BFI) using ResNet-101, which further verifies the robust plug-and-play ability of our FPT. The visualization results in instance segmentation are given in Fig.~\ref{fig:5}. Compared to other feature interaction methods, the results of FPT show more precise predictions for both small (\eg, bottle) and large objects (\eg, bicycle). Moreover, it shows the gracile parts in the object (\eg, the horse legs) are also well predicted using our FPT.
\begin{table*}[t]
\scriptsize
\begin{center}
\renewcommand\arraystretch{1.2}
\setlength{\tabcolsep}{1.0pt}{
\begin{tabular}{r|r|c:cc:cc:cc:cc:cc:c}
Methods & Backbone & \multicolumn{2}{c}{{AP}} & \multicolumn{2}{c}{{AP}$_{50}$} & \multicolumn{2}{c}{{AP}$_{75}$} & \multicolumn{2}{c}{{AP}$_S$} & \multicolumn{2}{c}{{AP}$_M$} & \multicolumn{2}{c}{{AP}$_L$} \\  \hline \hline
\multirow{4}{*}{BFP~\cite{lin2017feature}} & ResNet-101 & 33.1 & 32.6 & 53.8 & 51.7 & 34.6 & 33.3 & 12.6 & 11.4 & 35.3 & 34.4 & 49.5 & 48.9 \\ 
 & NL-ResNet-101 & 34.4 & 33.7 & 54.3 & 53.6 & 35.8 & 33.9 & 15.1 & 13.7 & 37.1 & 36.0 & 50.7 & 49.7 \\ 
 & GC-ResNet-101 & 35.0 & 34.2 & 55.8 & 54.1 & 36.5 & 35.3 & 14.8 & 13.9 & 38.6 & 37.3 & 50.9 & 50.5 \\ 
 & AA-ResNet-101 & 33.8 & 32.8 & 54.2 & 52.3 & 35.4 & 33.8 & 13.0 & 12.3 & 35.5 & 34.5 & 50.0 & 49.0 \\ \hline 
BFP+FPN~\cite{lin2017feature} & ResNet-101 & 36.2 & 35.7 & 59.1 & 58.0 & 39.0 & 37.8 & 18.2 & 15.5 & 39.0 & 38.1 & 52.4 & 49.2\\ 
BFP+BPA~\cite{liu2018path} & ResNet-101 & 37.3 & 36.3 & 60.4 & 59.0 & 39.9 & 38.3 & 18.9 & 16.3 & 39.7 & 39.0 & 53.0 & 50.5  \\ 
BFP+BFI~\cite{lin2019zigzagnet} & ResNet-101 & 39.5 & - & - & - & - & - & - & - & - & - & - & -  \\  \hline 
\multirow{4}{*}{BFP+FPT}~~& ResNet-101 & 41.6 & 38.6 & 60.9 & 58.2 & 44.0 & 43.3 & 23.4 & 19.0 & 41.5 & 39.2 & 53.1 & 50.8 \\ 
 & NL-ResNet-101 & 42.0 & 39.5 & 62.1 & 60.7 & 46.5 & 45.4 & 25.1 & 20.8 & 42.6 & 41.0 & 53.7 & 53.0 \\ 
 & GC-ResNet-101 & 42.5 & \textbf{40.3} & 62.0 & 61.0 & 46.1 & 45.8 & \textbf{25.3} & 21.1 & 42.7 & \textbf{41.8} & 53.1 & 52.7 \\ 
 & AA-ResNet-101 & 42.1 & 40.1 & 61.5 & 60.1 & 46.5 & 45.2 & 25.2 & 20.6 & 42.6 & 41.2 & 53.5 & 52.0 \\  \hline 
BFP+FPT [AH] & ResNet-101 & 41.1 & 40.0 & 62.0 & 59.9 & 46.6 & 45.5 & 24.2 & 20.5 & 42.1 & 41.0 & 53.3 & 52.5 \\ 
BFP+FPT [MT] & ResNet-101 & 41.2 & 39.8 & 62.1 & 60.1 & 46.0 & 45.1 & 24.1 & 20.9 & 41.9 & 40.8 & 53.2 & 51.9 \\ \hline  
BFP+FPN~\cite{lin2017feature} [all] & ResNet-101 & 37.9 & 36.3 & 59.6 & 58.8 & 40.1 & 39.1 & 19.5 & 16.7 & 41.0 & 40.3 & 53.5 & 51.1 \\ 
BFP+BPA~\cite{liu2018path} [all] & ResNet-101 & 39.0 & 37.7 & 60.8 & 59.4 & 41.7 & 40.1 & 20.2 & 18.5 & 41.5 & 40.1 & 54.1 & 52.4 \\ 
BFP+BFI~\cite{lin2019zigzagnet} [all] & ResNet-101 & 40.1 & 38.2 & 61.2 & 60.0 & 42.6 & 42.4 & 21.9 & 19.6 & 42.4 & 40.8 & 54.3 & 52.5  \\
BFP+FPT [all] & ResNet-101 & \textbf{42.6} & \textbf{40.3} & \textbf{62.4} & \textbf{61.1} & \textbf{46.9} & \textbf{45.9} & 24.9 & \textbf{21.3} & \textbf{43.0} & 41.2 & \textbf{54.5} & \textbf{53.3}
\end{tabular}}
\end{center}
\caption{Experimental results on MS-COCO 2017 \emph{test-dev}~\cite{lin2014microsoft}. ``AH'' is Augmented Head, and ``MT'' is Multi-scale Training~\cite{liu2018path}; ``all'' means that both the AH and MT are used. Results on the left and right of the dashed are of bounding box detection and instance segmentation. ``-'' means that there is no reported result in its paper.}
\label{tab2}
\end{table*}
\subsection{Experiments on Pixel-Level Recognition}
\myparagraph{Dataset.}
Our pixel-level segmentation experiments were conducted on four benchmarks:
(1)~\emph{Cityscapes}~\cite{cordts2016cityscapes} contains $19$ classes, and includes $2,975$, $500$ and $1,525$ images for training, validation and test, respectively;
(2)~\emph{ADE20K}~\cite{zhou2017scene} has $150$ classes, and uses $20k$, $2k$, and $3k$ images for training, validation and test, respectively;
(3)~\emph{LIP}~\cite{gong2017look} contains $50,462$ images with $20$ classes, and includes $30,462$, $10k$ and $10k$ images for training, validation and test, respectively;
(4)~\emph{PASCAL VOC 2012}~\cite{everingham2015pascal} contains $21$ classes, and includes $1,464$, $1,449$ and $1,456$ images for training, validation and test, respectively.

\myparagraph{Backbone.}
We used dilated ResNet-101~\cite{yu2015multi} as the backbone as in~\cite{zhang2019co}.

\myparagraph{Setting.}
We first pre-trained the backbone network on the ImageNet~\cite{deng2009imagenet}, then fine-tuned the whole network on the training data while fixing the parameters of backbone as in~\cite{zhang2018context}. Before input, we cropped the image into $969\times 969$ for \emph{Cityscapes}, $573\times 573$ for \emph{LIP}, and $521\times 521$ for \emph{PASCAL VOC 2012}. Because images in ADE20K are of various sizes, we cropped the shorter-edge images to an uniform size $\{269, 369, 469, 569\}$ as that in~\cite{yuan2018ocnet}.

\begin{table}[t]
\scriptsize
\begin{center}
\renewcommand\arraystretch{1.2}
\setlength{\tabcolsep}{2.0pt}{
\begin{tabular}{r|cc|rr}
Methods~~& Tra.mIoU & Val.mIoU & Params & GFLOPs  \\  \hline \hline
UFP~\cite{peng2017large}~~& {86.0} & {79.1} & 71.3 M  & 916.1 \\ \hline
UFP+ST~\cite{peng2017large}~~& 86.9 & 80.7& 91.2 M & 948.4 \\
UFP+LGT~\cite{peng2017large}~~& 86.5 & 80.3 & 102.8 M & 1008.3 \\
UFP+RT~\cite{peng2017large}~~& 86.3 & 80.1 & 77.4 M & 929.3 \\ \hline
UFP+LGT+ST~\cite{peng2017large}~~& 87.2 & 80.9 & 121.3 M & 1052.6 \\
UFP+RT+ST~\cite{peng2017large}~~& 87.0 & 80.8 & 96.2 M & 985.2 \\
UFP+LGT+RT~\cite{peng2017large}~~& 86.6 & 80.4 & 107.0 M & 1014.8 \\
UFP+LGT+ST+RT~\cite{peng2017large}~~& {\textbf{87.4}} & {\textbf{81.7}} & 127.2 M & 1063.9 \\  \hline
the improvement~~& $\uparrow$~\textbf{1.4} & $\uparrow$~\textbf{2.6} & ~~ & ~~
\end{tabular}}
\end{center}
\caption{Ablation study on the Cityscapes validation set~\cite{cordts2016cityscapes}. ``LGT'' is Locality-constrained Grounding Transformer; ``RT'' is Rendering Transformer; ``ST'' is Self-Transformer. ``+'' means building the method on the top of UFP.}
\label{tab3}
\end{table}
\begin{table*}[t]
\scriptsize
\begin{center}
\renewcommand\arraystretch{1.2}
\setlength{\tabcolsep}{0.5pt}{
\begin{tabular}{r|c|cccc}
{Methods}~ & {Backbone}  & {Cityscapes} & ADE20K & LIP & {PASCAL VOC 2012} \\ \hline\hline
baseline~ & ResNet-101 & 65.3 & 40.9 & 42.7 & 62.2 \\ \hline
CFNet~\cite{zhang2019co}~ & ResNet-101& 80.6 & 44.9 & 54.6 & 84.2 \\
AFNB~\cite{zhu2019asymmetric}~ & ResNet-101 & 81.3 & 45.2 & - & - \\
HRNet~\cite{sun2019high}~ & HRNetV2-W48 & 81.6 & 44.7 & {\color{blue}\textbf{55.9}} & {\color{blue}\textbf{84.5}}  \\
OCNet~\cite{yuan2018ocnet}~ & ResNet-101 & 81.7 & {\color{blue}\textbf{45.5}} & 54.7 & 84.3 \\
GSCNN~\cite{takikawa2019gated}~ & Wide-ResNet-101 & {\color{red}\textbf{82.8}} & - & 55.2 & - \\  \hline
PPM~\cite{zhao2017pyramid}+OC~\cite{yuan2018ocnet}~ & ResNet-101 & 79.9 & 43.7 & 53.0 & 82.9  \\
ASPP~\cite{chen2017rethinking}+OC~\cite{yuan2018ocnet}~ & ResNet-101 & 80.0 & 44.1 & 53.3 & 82.7 \\
UFP~\cite{peng2017large}+OC~\cite{yuan2018ocnet}~ & ResNet-101 & 80.6 & 44.7 & 54.5 & 83.2  \\ \hline
PPM~\cite{zhao2017pyramid}+FPT~ & ResNet-101~ & 80.4($\uparrow$~\textbf{0.5}) & 44.8($\uparrow$~\textbf{1.1}) & 54.2($\uparrow$~\textbf{1.2}) & 83.2($\uparrow$~\textbf{0.3})  \\
ASPP~\cite{chen2017rethinking}+FPT~ & ResNet-101~& 80.7($\uparrow$~\textbf{0.7}) & 45.2($\uparrow$~\textbf{1.1}) & 54.4($\uparrow$~\textbf{1.1}) & 83.1($\uparrow$~\textbf{0.4}) \\
UFP~\cite{peng2017large}+FPT~ & ResNet-101 & {\color{blue}\textbf{82.2}}($\uparrow$~\textbf{1.6}) & {\color{red}\textbf{45.9}}($\uparrow$~\textbf{1.2}) & {\color{red}\textbf{56.2}}($\uparrow$~\textbf{1.7}) & {\color{red}\textbf{85.0}}($\uparrow$~\textbf{1.8}) \\
\end{tabular}}
\end{center}
\caption{Comparisons with state-of-the-art on test sets of Cityscapes~\cite{cordts2016cityscapes} and PASCAL VOC 2012~\cite{everingham2015pascal}, validation sets of ADE20K~\cite{zhou2017scene} and LIP~\cite{gong2017look}. Results in this table refer to mIoU; ``-'' means that there is no reported result in its paper. The {\color{red}\textbf{best}} and {\color{blue}\textbf{second best}} models under each setting are marked with corresponding formats.}
\label{tab4}
\end{table*}
\myparagraph{Training details.}
We followed~\cite{yuan2018ocnet} to use the learning rate scheduling $lr=baselr\times(1-\frac{iter}{total_{iter}})^{power}$. On \emph{Cityscapes}, \emph{LIP} and \emph{PASCAL VOC 2012}, the base learning rate was $0.01$, and the power is $0.9$. The weight decay and momentum were set to $0.0005$ and $0.9$, respectively. On \emph{ADE20K}, the base learning rate was $0.02$ and the power was $0.9$. The weight decay and momentum were $0.0001$ and $0.9$, respectively. We trained models on $8$ GPUs with SBN~\cite{zhang2018context}. The model was trained for $120$ epochs on \emph{Cityscapes} and \emph{ADE20K}, $50$ on \emph{LIP}, and $80$ on \emph{PASCAL VOC 2012}. For data augmentation, the training images were flipped left-right and randomly scaled between a half and twice as in~\cite{zhang2019co}.

\myparagraph{Comparison methods.}
Our FPT was applied to the feature pyramids constructed by three methods: UFP~\cite{peng2017large}, PPM~\cite{he2015spatial,zhao2017pyramid} and ASPP~\cite{chen2017rethinking}. Based on each of these methods, we compared our FPT to the state-of-the-art pixel-level feature pyramid interaction method, \ie, Object Context Network (OCNet)~\cite{yuan2018ocnet}.

\myparagraph{Metrics.}
We used the standard mean Intersection of Union (mIoU) as a uniform metric. We showed the results of ablation study by reporting the mIoU of training set (\ie, Tra.mIoU) and validation set (\ie, Val.mIoU) on the \emph{Cityscapes}.

\myparagraph{Ablation study.}
Results are given in Table~\ref{tab3}.
UFP is the baseline. Applying our transformers (\ie, +ST, +LGT and +RT) to UFP respectively achieves the improvements of $0.9\%$, $0.5\%$ and $0.3\%$ Tr.mIoU, and the more impressive $1.6\%$, $1.2\%$ and $1.0\%$ Val.mIoU. Moreover, any component combinations of our transformers yields concretely better results than using individual ones. Our best model achieves $1.4\%$ and $2.6\%$ improvements (over UFP) for Tr.mIoU and Val.mIoU, respectively.

\myparagraph{Model efficiency.}
In Table~\ref{tab3}, we reported the model Params and GFLOPs. It is clear that using our transformers increases a fair computational overhead. For example, +ST, +LGT and +RT respectively add Params $0.28 \times$, $0.44 \times$ and $0.09 \times$, and increase GFLOPs by $0.04 \times$, $0.10 \times$ and $0.01 \times$, compared to UFP.

\begin{figure}[t]
\centering
\includegraphics[width=.9 \textwidth]{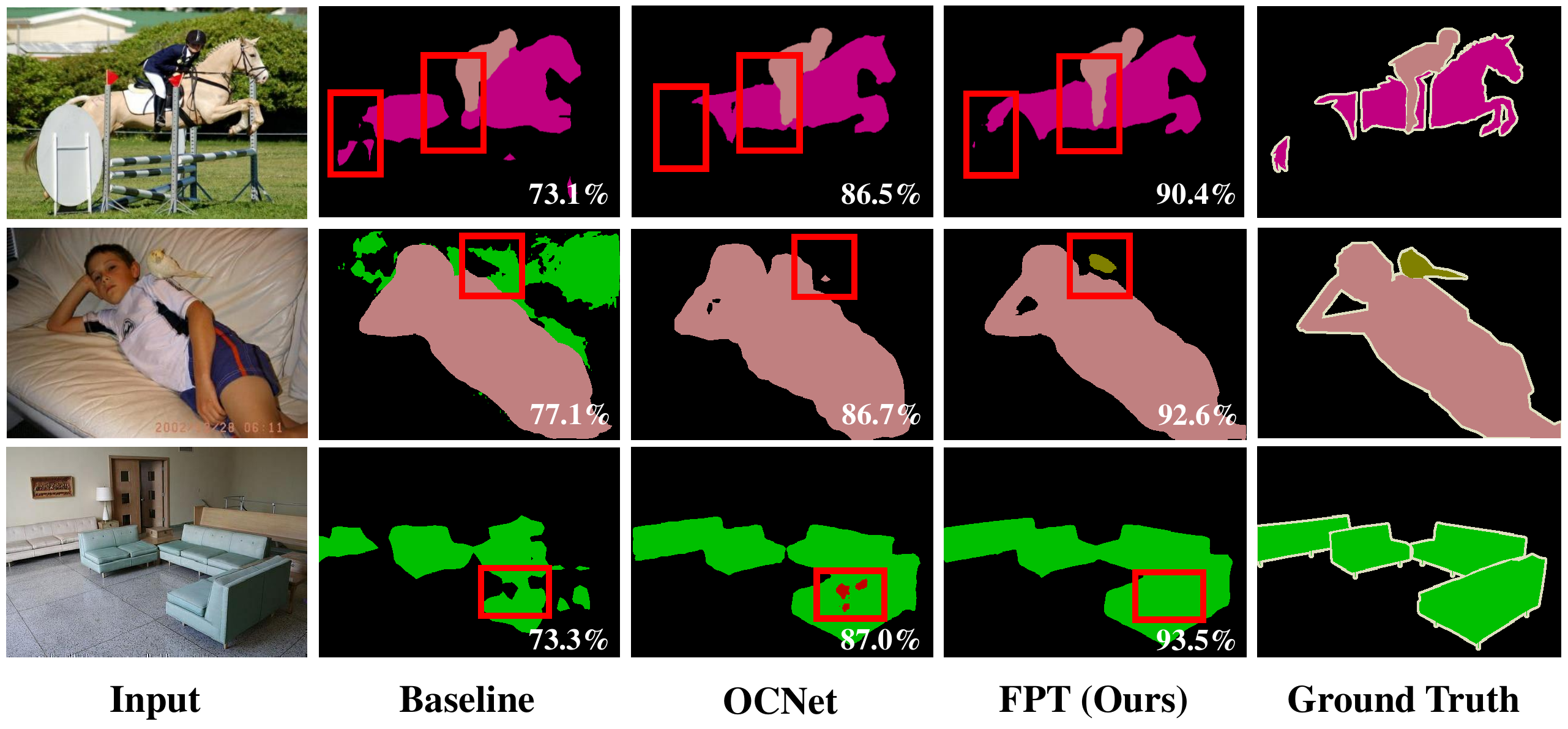}
\caption{Visualization results in segmentation. The red rectangle highlights the better predicted area of FPT. Samples are from the validation set of \emph{PASCAL VOC 2012} ~\cite{everingham2015pascal}. The value on each image represents the corresponding segmentation mIoU.}
\label{fig:6}
\end{figure}

\myparagraph{Comparing to the state-of-the-arts.}
From Table~\ref{tab4}, we can observe that our FPT can achieve a new state-of-the-art performance over all the previous methods based on the same backbone (\ie, ResNet-101). It obtains impressive improvements as $1.6\%$, $1.2\%$, $1.7\%$ and $1.8\%$ mIoU on \emph{Cityscapes}~\cite{cordts2016cityscapes}, \emph{ADE20K}~\cite{zhou2017scene}, \emph{LIP}~\cite{gong2017look} and \emph{PASCAL VOC 2012}~\cite{everingham2015pascal}, respectively. Besides, compared to OCNet, FPT obtains gain by 0.9\%, 1.1\%, 1.3\% and 0.8\% mIoU in these four datasets on average. In Fig.~\ref{fig:6}, we provide the qualitative results of our method\footnote{More visualization results are given in the \emph{Section E} of the supplementary.}. Compared to the baseline~\cite{peng2017large} and OCNet~\cite{yuan2018ocnet}, the results of FPT show more precise segmentation on boundaries, particularly for smaller and thinner objects, \eg, the guardrail, person's leg and bird. Moreover, FPT can also achieve more integrated segmentation on some larger objects, \eg, the horse, person and sofa.
\section{Conclusion}
We proposed an efficient feature interaction approach called FPT, composed of three carefully-designed transformers to respectively encode the explicit self-level, top-down and bottom-up information in the feature pyramid. Our FPT does not change the size of the feature pyramid, and is thus \emph{generic} and easy to plug-and-play with modern deep networks. Our extensive quantitative and qualitative results on three challenging visual recognition tasks showed that FPT achieves consistent improvements over the baselines and the state-of-the-arts, validating its high effectiveness and strong application capability.
\section*{Acknowledgements}
We would like to thank all the anonymous reviewers for their constructive comments. This work was partially supported by the National Key Research and Development Program of China under Grant 2018AAA0102002, the National Natural Science Foundation of China under Grant 61925204, the China Scholarships Council under Grant 201806840058, the Singapore Ministry of Education (MOE) Academic Research Fund (AcRF) Tier 1 grant, and the NTU-Alibaba JRI.
\bibliographystyle{splncs}
\bibliography{6141.bbl}
\clearpage
\section*{Supplementary Materials}
\setcounter{table}{0}
\renewcommand{\thetable}{S\arabic{table}}
\setcounter{figure}{0}
\renewcommand{\thefigure}{S\arabic{figure}}
\setcounter{section}{0}
\renewcommand\thesection{\Alph{section}}
These materials include the details of the effectiveness of $F_{eud}$ (Section \ref{sec:s1}),
an additional study on hyperparameters (Section \ref{sec:s2}),
FPT complexity analysis (Section \ref{sec:s3}),
more quantitative result comparisons (Section \ref{sec:s4}),
and more qualitative results (Section \ref{sec:s5}).
\section{Effectiveness of $F_{eud}$}\label{sec:s1}
In Section 3.3 of the main paper, we propose to use the negative value of euclidean distance  $F_{eud}$~\cite{zhang2018learning} (instead of the conventional $F_{sim}$~\cite{vaswani2017attention}) as the similarity function in Grounding Transformer (GT).
In this section, we show the effectiveness of $F_{eud}$.
In details, the Mixture of Softmaxes (MoS)~\cite{yang2017breaking} is deployed as the normalizing function, and the number of the divided parts $\mathcal{N}$ is set to $4$.
Table~\ref{tab:1} shows that $F_{eud}$ surpasses the classic \emph{softmax}-based $F_{sim}$ in both cases of GT (with and without MoS).
In particular, $F_{eud}$ with MoS achieves the best performance. There are at most $3.1\%$ box AP and $3.3\%$ mask AP improvements for object detection and instance segmentation, respectively.
\begin{table*}
\scriptsize
\begin{center}
\renewcommand\arraystretch{1.3}
\setlength{\tabcolsep}{2.5pt}{
\begin{tabular}{c|c:cc:cc:cc:cc:cc:c}
Methods & \multicolumn{2}{c}{AP} & \multicolumn{2}{c}{AP$_{50}$} & \multicolumn{2}{c}{AP$_{75}$} & \multicolumn{2}{c}{AP$_S$} & \multicolumn{2}{c}{AP$_M$} & \multicolumn{2}{c}{AP$_L$}\\ \hline \hline
BFP~\cite{lin2017feature} & 31.6 & 29.9 & 54.1 & 50.7 & 35.9 & 34.7 & 16.1 & 14.2 & 32.5 & 31.6 & 48.8 & 48.5\\ \hline
+ $F_{sim}$~\cite{vaswani2017attention} & 32.2 & 30.5 & 54.2 & 50.9 & 35.6 & 34.5 & 16.3 & 14.5 & 32.1 & 31.2 & 49.0 & 48.6 \\
+ $F_{eud}$~\cite{zhang2018learning} & 33.7 & 32.1 & 54.5 & 52.1 & 36.0 & 34.9 & 16.9 & 15.6 & 33.0 & 31.8 & 49.4 & 48.7  \\
+ $F_{sim}$~\cite{vaswani2017attention} + MoS~\cite{yang2017breaking} & 32.6 & 31.1 & 54.3 & 51.2 & 35.8 & 34.8 & 16.4 & 15.3 & 32.6 & 31.5 & 49.1 & 48.5 \\
+ $F_{eud}$~\cite{zhang2018learning} + MoS~\cite{yang2017breaking} & \textbf{34.7} & \textbf{33.2} & \textbf{55.4} & \textbf{53.2} & \textbf{37.0} & \textbf{36.1} & \textbf{17.8} & \textbf{16.2} & \textbf{33.9} & \textbf{32.0} & \textbf{50.3} & \textbf{49.1} \\
\end{tabular}}
\end{center}
\caption{Comparing $F_{eud}$ with $F_{sim}$ on validation set of MS-COCO 2017~\cite{lin2014microsoft}. The backbone is ResNet-50~\cite{he2016deep}. ``BFP'' is the bottom-up feature pyramid (BFP)~\cite{lin2017feature}. Results on the left and right of the dashed are respectively from bounding box detection and instance segmentation.}
\label{tab:1}
\end{table*}
\section{Hyperparameters}\label{sec:s2}
\subsection{$\mathcal{N}$ in ST}
In Section 3.2, we use MoS~\cite{yang2017breaking} as the normalizing function. In  this section, we investigate the influence of $\mathcal{N}$ (in MoS) on Self-Transformer (ST).
In particular, no MoS~\cite{yang2017breaking} means $\mathcal{N}$=1, \ie, the classical \emph{softmax}~\cite{wang2018non}.
In Table~\ref{tab:4}, we can see that $\mathcal{N}$=2 brings the best performance in all cases.
\begin{table*}[t]
\scriptsize
\begin{center}
\renewcommand\arraystretch{1.3}
\setlength{\tabcolsep}{2.5pt}{
\begin{tabular}{c|c:cc:cc:cc:cc:cc:c}
$\mathcal{N}$ & \multicolumn{2}{c}{AP} & \multicolumn{2}{c}{AP$_{50}$} & \multicolumn{2}{c}{AP$_{75}$} & \multicolumn{2}{c}{AP$_S$} & \multicolumn{2}{c}{AP$_M$} & \multicolumn{2}{c}{AP$_L$}\\ \hline \hline
BFP~\cite{lin2017feature} & 31.6 & 29.9 & 54.1 & 50.7 & 35.9 & 34.7 & 16.1 & 14.2 & 32.5 & 31.6 & 48.8 & 48.5\\ \hline
1 (w/o MoS~\cite{yang2017breaking}) & 31.7 & 30.0 & 54.3 & 50.5 & 36.0 & 34.9 & 16.3 & 14.3 & 32.6 & 31.9 & 48.5 & 48.4 \\
2 & \textbf{31.8} & \textbf{30.2} & \textbf{54.6} & \textbf{51.1} & \textbf{36.5} & \textbf{35.1} & \textbf{16.8} & \textbf{14.6} & \textbf{33.2} & \textbf{32.0} & \textbf{49.8} & \textbf{49.3} \\
4 & 31.1 & 29.3 & 54.0 & 50.5 & 35.9 & 34.6 & 16.4 & 14.1 & 32.8 & 31.4 & 49.1 & 48.3 \\
6 & 30.6 & 28.7 & 53.6 & 49.7 & 35.7 & 34.0 & 16.1 & 13.7 & 32.2 & 30.9 & 48.8 & 48.0 \\
8 & 30.0 & 28.1 & 53.1 & 49.5 & 35.3 & 33.7 & 15.9 & 13.3 & 31.8 & 30.5 & 48.2 & 47.5 \\
\end{tabular}}
\end{center}
\caption{The influence of $\mathcal{N}$ on ST. Experiments are carried out on validation set of MS-COCO 2017~\cite{lin2014microsoft}. The backbone is ResNet-50~\cite{he2016deep}. ``BFP'' is the bottom-up feature pyramid (BFP)~\cite{lin2017feature}. ``w/o MoS'' means that these results are obtained without MoS~\cite{yang2017breaking}. Results on the left and right of the dashed are of bounding box detection and instance segmentation.}
\label{tab:4}
\end{table*}
\subsection{$\mathcal{N}$ in GT}
In this section, we investigate the influence of $\mathcal{N}$ (in MoS) on GT.
As shown in Table~\ref{tab:2}, we can see that using $\mathcal{N}$=4 achieves the best performance for both object detection and instance segmentation.
\begin{table*}[t]
\scriptsize
\begin{center}
\renewcommand\arraystretch{1.3}
\setlength{\tabcolsep}{2.5pt}{
\begin{tabular}{c|c:cc:cc:cc:cc:cc:c}
$\mathcal{N}$ & \multicolumn{2}{c}{AP} & \multicolumn{2}{c}{AP$_{50}$} & \multicolumn{2}{c}{AP$_{75}$} & \multicolumn{2}{c}{AP$_S$} & \multicolumn{2}{c}{AP$_M$} & \multicolumn{2}{c}{AP$_L$}\\ \hline \hline
BFP~\cite{lin2017feature} & 31.6 & 29.9 & 54.1 & 50.7 & 35.9 & 34.7 & 16.1 & 14.2 & 32.5 & 31.6 & 48.8 & 48.5\\ \hline
1 (w/o MoS~\cite{yang2017breaking}) & 33.7 & 32.1 & 54.5 & 52.1 & 36.0 & 34.9 & 16.9 & 15.6 & 33.0 & 31.8 & 49.4 & 48.6 \\
2 & 34.3 & 32.7 & 55.1 & 52.8 & 36.5 & 35.4 & 17.3 & 15.9 & 33.5 & 31.6 & 49.9 & 48.8 \\
4 & \textbf{34.7} & \textbf{33.2} & \textbf{55.4} & \textbf{53.2} & \textbf{37.0} & \textbf{36.1} & \textbf{17.8} & \textbf{16.2} & \textbf{33.9} & \textbf{32.0} & \textbf{50.3} & \textbf{49.1} \\
6 & 33.4 & 32.9 & 54.3 & 52.7 & 36.4 & 35.6 & 17.3 & 15.7 & 33.5 & 31.5 & 50.0 & 48.7 \\
8 & 32.5 & 32.3 & 53.3 & 52.0 & 35.9 & 35.0 & 17.0 & 15.2 & 32.9 & 30.7 & 49.5 & 48.4 \\
\end{tabular}}
\end{center}
\caption{The influence of $\mathcal{N}$ on GT. Experiments are carried out on validation set of MS-COCO 2017~\cite{lin2014microsoft}. The backbone is ResNet-50~\cite{he2016deep}. ``BFP'' is the bottom-up feature pyramid (BFP)~\cite{lin2017feature}. ``w/o MoS'' means that these results are obtained without MoS~\cite{yang2017breaking}. Results on the left and right of the dashed are of bounding box detection and instance segmentation.}
\label{tab:2}
\end{table*}
\subsection{$\emph{square\_size}$ in LGT}
In Section 3.3, we introduce LGT for semantic segmentation. In this section, we investigate the influence of the side length $\emph{square\_size}$ (of local square area) on LGT. We use MoS~\cite{yang2017breaking} with $\mathcal{N}$=4 as the normalizing function.
We report the standard mean Intersection of Union (mIoU) on the training set (\ie, Tra.mIoU) as well as the validation set (\ie, Val.mIoU) of Cityscapes~\cite{cordts2016cityscapes}, in Table~\ref{tab:3}. We can see that LGT with $square\_size$=5 achieves the best performance.
\begin{table*}[t]
\scriptsize
\begin{center}
\renewcommand\arraystretch{1.3}
\setlength{\tabcolsep}{2.5pt}{
\begin{tabular}{c|ccc}
$square\_size$ & Method & Tra.mIoU~(\%) & Val.mIoU~(\%)\\  \hline \hline
- & backbone + UFP & 86.0 & 79.1 \\ \hline
1 & backbone + UFP + LGT & 86.1 & 79.5 \\
3 & backbone + UFP + LGT & 86.2 & 79.9 \\
5 & backbone + UFP + LGT & \textbf{86.3} & \textbf{80.0} \\
7 & backbone + UFP + LGT & 86.1 & 79.8 \\
9 & backbone + UFP + LGT & 85.8 & 79.6 \\
\end{tabular}}
\end{center}
\caption{The influence of $\emph{square\_size}$ of LGT on the pixel-level semantic segmentation task. The backbone is the dilated ResNet-101~\cite{yu2015multi}. Experiments are carried out on training set and the validation set of Cityscapes~\cite{cordts2016cityscapes}. ``UFP'' is the unscathed feature pyramid.}
\label{tab:3}
\end{table*}
\subsection{DropBlock in Instance-level Tasks}
In Section 3.5, we apply the DropBlock~\cite{ghiasi2018dropblock} to each transformed feature map, to alleviate the over-fitting problem. In this section, we investigate the influence of two hyper-parameters (\ie, the drop block size $block\_size$ and the keep probability $keep\_prob$ of each feature position) of DropBlock~\cite{ghiasi2018dropblock} on instance-level tasks (\ie, object detection and instance segmentation). The MoS~\cite{yang2017breaking} is applied in GT with its $\mathcal{N}$=4, and in ST with its $\mathcal{N}$=2. In Table~\ref{tab:5}, we find that $block\_size$=5 and $keep\_prob$=0.9 result the best performance.
\begin{table*}[t]
\scriptsize
\begin{center}
\renewcommand\arraystretch{1.3}
\setlength{\tabcolsep}{2.5pt}{
\begin{tabular}{c|c:cc:cc:cc:c}
Settings & \multicolumn{2}{c}{$block\_size$=1} & \multicolumn{2}{c}{$block\_size$=3} & \multicolumn{2}{c}{$block\_size$=5} & \multicolumn{2}{c}{$block\_size$=7}\\ \hline \hline
$keep\_prob$=0.1 & 30.7 & 29.7 & 31.4 & 30.7 & 31.9 & 30.1 & 30.0 & 29.8 \\
$keep\_prob$=0.3 & 32.1 & 30.9 & 32.9 & 31.6 & 33.2 & 31.1 & 31.8 & 30.8 \\
$keep\_prob$=0.5 & 33.2 & 31.0 & 34.2 & 33.5 & 35.5 & 34.7 & 33.7 & 33.4 \\
$keep\_prob$=0.7 & 33.8 & 32.4 & 35.6 & 34.9 & 36.1 & 35.7 & 35.8 & 34.6 \\
$keep\_prob$=0.9 & 34.2 & 33.8 & 36.6 & 35.1 & \textbf{38.0} & \textbf{36.8} & 36.6 & 35.3 \\
\end{tabular}}
\end{center}
\caption{The influence of $block\_size$ and $keep\_prob$ of DropBlock~\cite{ghiasi2018dropblock} on the instance-level tasks (\ie, object detection and instance segmentation). The backbone is ResNet-50~\cite{he2016deep}. Results on the left and right of the dashed are AP of bounding box detection and mask AP of instance segmentation.}
\label{tab:5}
\end{table*}
\subsection{DropBlock in Pixel-level Task}
In this section, we investigate the influence of two hyper-parameters (\ie, the drop block size $block\_size$ and the keep probability $keep\_prob$ of each feature position) of DropBlock~\cite{ghiasi2018dropblock} on the pixel-level semantic segmentation. The MoS~\cite{yang2017breaking} is applied in GT with its $\mathcal{N}$=4, and in ST with its $\mathcal{N}$=2.
The $\emph{square\_size}$ of LGT is set to $5$.
We report mIoU on the validation set (\ie, Val.mIoU) of Cityscapes~\cite{cordts2016cityscapes} in Table~\ref{tab:6}. We find that using $block\_size$=3 and $keep\_prob$=0.9 achieves the best performance.
\begin{table*}[t]
\scriptsize
\begin{center}
\renewcommand\arraystretch{1.3}
\setlength{\tabcolsep}{2.5pt}{
\begin{tabular}{c|cccc}
Settings & {$block\_size$=1} & {$block\_size$=3} & {$block\_size$=5} & {$block\_size$=7}\\ \hline \hline
$keep\_prob$=0.1 & 80.8 & 81.0 & 80.8 & 80.4 \\
$keep\_prob$=0.3 & 81.0 & 81.1 & 81.0 & 80.7 \\
$keep\_prob$=0.5 & 81.2 & 81.4 & 81.2 & 80.9 \\
$keep\_prob$=0.7 & 81.4 & 81.6 & 81.3 & 81.1 \\
$keep\_prob$=0.9 & 81.3 & \textbf{81.7} & 81.5 & 81.4 \\
\end{tabular}}
\end{center}
\caption{The influence of $block\_size$ and $keep\_prob$ of DropBlock~\cite{ghiasi2018dropblock} on the pixel-level semantic segmentation. The backbone is the dilated ResNet-101~\cite{yu2015multi}. Experiments are carried out on validation set of Cityscapes~\cite{cordts2016cityscapes}. Results in this table refer to the mIoU on the validation set (\ie, Val.mIoU).}
\label{tab:6}
\end{table*}
\section{FPT Complexity}\label{sec:s3}
In Section 4.1.1, we report the model efficiency. In this section, we supplement the details of model Parameters (Params) and FLOPs using Mask R-CNN~\cite{he2017mask} head.
In Table~\ref{tab:7}, we compare our FPT and its components (\ie, ST, GR, and RT) to the non-local operation on the validation set of MS-COCO 2017 for instance segmentation~\cite{lin2014microsoft}.
The implementation detail of the non-local operation is the same as that in~\cite{wang2018non}.

From Table~\ref{tab:7}, we can find that for the non-local operation the average increases of the model Params and FLOPs required by AP at per improved point are $0.27\times$ and $0.27\times$, respectively.
In contrast, the average increases in our FPT are lower (better) as $0.21\times$ and $0.15\times$, respectively.
\begin{table*}[t]
\tiny
\begin{center}
\renewcommand\arraystretch{1.3}
\setlength{\tabcolsep}{1pt}{
\begin{tabular}{c|cccccc:cc}
Methods & {AP} & {AP$_{50}$} & {AP$_{75}$} & {AP$_S$} & {AP$_M$} & {AP$_L$} & Params & FLOPs\\ \hline \hline
BFP~\cite{lin2017feature} & 29.9 & 50.7 & 34.7 & 14.2 & 31.6 & 48.5 & $1\times$ & $1\times$ \\ \hline
+ non-local~\cite{wang2018non} & 30.8 ($\uparrow$~\textbf{0.9}) & 52.4 ($\uparrow$~\textbf{1.7}) & 35.5 ($\uparrow$~\textbf{0.8}) & 15.2 ($\uparrow$~\textbf{1.0}) & 32.5 ($\uparrow$~\textbf{0.9}) & 49.5 ($\uparrow$~\textbf{1.0}) & $1.24\times$ & $1.24\times$ \\ \hline
+ ST & 30.6 ($\uparrow$~\textbf{0.7}) & 51.4 ($\uparrow$~\textbf{0.7}) & 35.5 ($\uparrow$~\textbf{0.8}) & 15.1 ($\uparrow$~\textbf{0.9}) & 32.1 ($\uparrow$~\textbf{0.5}) & 49.7 ($\uparrow$~\textbf{1.2}) & $1.59\times$ & $1.44\times$ \\
+ GT & 33.9 ($\uparrow$~\textbf{4.0}) & 52.4 ($\uparrow$~\textbf{1.7}) & 37.7 ($\uparrow$~\textbf{3.0}) & 16.9 ($\uparrow$~\textbf{2.7}) & 33.3 ($\uparrow$~\textbf{1.7}) & 51.7 ($\uparrow$~\textbf{3.2}) &  $1.85\times$ & $1.54\times$ \\
+ RT & 33.1 ($\uparrow$~\textbf{3.2}) & 52.0 ($\uparrow$~\textbf{1.3}) & 37.7 ($\uparrow$~\textbf{3.0}) & 15.3 ($\uparrow$~\textbf{1.1}) & 34.9 ($\uparrow$~\textbf{3.3}) & 52.1 ($\uparrow$~\textbf{3.6}) &  $1.15\times$ & $1.09\times$ \\ \hline
+ FPT & 36.8 ($\uparrow$~\textbf{6.9}) & 55.9 ($\uparrow$~\textbf{5.2}) & 38.6 ($\uparrow$~\textbf{3.9}) & 18.8 ($\uparrow$~\textbf{4.6}) & 35.3 ($\uparrow$~\textbf{3.7}) & 54.2 ($\uparrow$~\textbf{5.7}) &  $2.54\times$ & $2.01\times$ \\
\end{tabular}}
\end{center}
\caption{Model complexity analysis on validation set of MS-COCO 2017~\cite{lin2014microsoft} for instance segmentation. The backbone is ResNet-50~\cite{he2016deep}. ``BFP'' is the bottom-up feature pyramid (BFP)~\cite{lin2017feature}.}
\label{tab:7}
\end{table*}
\section{More Quantitative Result Comparisons}\label{sec:s4}
\subsection{Results on Stronger Backbones}
In addition to ResNet~\cite{he2016deep}, we also employ the Non-local ResNet~\cite{wang2018non}, the Global Context Network (GC-ResNet)~\cite{cao2019gcnet}, and the Attention Augmented Network (AA-ResNet)~\cite{IrwanBello2019aanet} as backbone networks in the instance-level recognition. In this section, we report more quantitative results on these backbones in Table~\ref{tab:r2}.

In Table~\ref{tab:r2}, we can observe that BFP+FPT still achieves the better performance than BFP+FPN, BFP+BPA and BFP+BFI on the stronger backbone networks (\ie, NL-, GC-, and AA-ResNet). In particular, BFP+FPT achieves up to 40.8\% bounding box AP (and 38.7\% mask AP), while BFP+FPN, BFP+BPA and BFP+BFI can achieve 37.9\% bounding box AP (and 36.8\% mask AP), 38.8\% bounding box AP (and 37.6\% mask AP), and 39.0\% bounding box AP (and 37.9\% mask AP), respectively.
\begin{table*}[t]
\scriptsize
\begin{center}
\renewcommand\arraystretch{1.3}
\setlength{\tabcolsep}{2.5pt}{
\begin{tabular}{cc|c:cc:cc:cc:cc:cc:c}
Methods & Backbone & \multicolumn{2}{c}{AP} & \multicolumn{2}{c}{AP$_{50}$} & \multicolumn{2}{c}{AP$_{75}$} & \multicolumn{2}{c}{AP$_S$} & \multicolumn{2}{c}{AP$_M$} & \multicolumn{2}{c}{AP$_L$}\\ \hline \hline
BFP+FPN~\cite{lin2017feature} & NL-ResNet & 37.2 & 36.4 & 60.1 & 59.2 & 40.0 & 38.5 & 19.0 & 16.7 & 37.8 & 37.1 & 51.1 & 49.9 \\
BFP+BPA~\cite{liu2018path} & NL-ResNet & 38.5 & 37.6 & 60.9 & 59.5 & 41.6 & 39.2 & 20.5 & 18.1 & 39.5 & 38.7 & 51.9 & 51.0 \\
BFP+BFI~\cite{lin2019zigzagnet} & NL-ResNet & 38.9 & 37.8 & 61.2 & 59.7 & 41.5 & 39.5 & 20.2 & 18.6 & 39.7 & 38.9 & 51.5 & 50.5 \\
BFP+FPT & NL-ResNet & 40.1 & 38.0 & 62.9 & 60.7 & 42.0 & 40.6 & 21.4 & 19.1 & 40.8 & 39.9 & 53.0 & 51.8 \\ \hline
BFP+FPN~\cite{lin2017feature} & GC-ResNet & 37.7 & 36.8 & 60.4 & 59.5 & 40.1 & 38.8 & 19.2 & 17.2 & 38.5 & 37.5 & 51.3 & 50.5 \\
BFP+BPA~\cite{liu2018path} & GC-ResNet & 38.8 & 37.4 & 61.2 & 59.8 & 41.9 & 40.3 & 20.8 & 18.5 & 39.7 & 38.9 & 52.2 & 51.5 \\
BFP+BFI~\cite{lin2019zigzagnet} & GC-ResNet & 39.0 & 37.7 & 62.0 & 60.2 & 42.3 & 40.7 & 21.1 & 18.9 & 40.2 & 39.1 & 52.0 & 51.8 \\
BFP+FPT & GC-ResNet & 40.4 & 38.5 & 63.3 & 61.0 & 43.5 & 41.9 & 22.6 & 19.7 & 41.1 & 40.5 & 53.4 & 52.3 \\ \hline
BFP+FPN~\cite{lin2017feature} & AA-ResNet & 37.9 & 36.7 & 60.7 & 59.6 & 40.3 & 38.4 & 19.6 & 17.5 & 38.6 & 37.3 & 51.8 & 50.1 \\
BFP+BPA~\cite{liu2018path} & AA-ResNet & 38.5 & 37.5 & 61.7 & 59.3 & 41.8 & 40.1 & 20.4 & 18.2 & 39.8 & 38.5 & 52.7 & 51.7 \\
BFP+BFI~\cite{lin2019zigzagnet} & AA-ResNet & 38.9 & 37.9 & 62.1 & 60.1 & 42.3 & 40.7 & 21.0 & 18.5 & 39.5 & 38.9 & 52.2 & 51.3 \\
BFP+FPT & AA-ResNet & 40.8 & 38.7 & 63.8 & 61.3 & 43.7 & 41.5 & 22.7 & 19.4 & 41.5 & 40.8 & 53.3 & 52.0 \\
\end{tabular}}
\end{center}
\caption{Combining FPN/BPA/BFI and NL-ResNet/GC-ResNet/AA-ResNet on validation set of MS-COCO 2017~\cite{lin2014microsoft}. The base is ResNet-50. Results on the left and right of the dashed are respectively from bounding box detection and instance segmentation.}
\label{tab:r2}
\end{table*}
\subsection{Results on Deeper Backbones}
In this section, we report more result comparisons on the deeper backbone network (\ie, ResNet-152) in Table~\ref{tab:r3}. We can observe that BFP + FPT on \textbf{ResNet-50}
achieves 38.0\%/36.8\% AP, which surpasses BFP + FPN~\cite{lin2017feature} and BFP + BPA~\cite{liu2018path} (\ie, 36.2\%/35.7\% AP and 37.3\% /36.3\% AP) on \textbf{ResNet-101} under the similar number of parameters (88 M). Compared to results on \textbf{ResNet-152}, FPT can still surpass BFP (35.8\%/34.6\% AP) with fewer parameters. Although BFP+FPN on \textbf{ResNet-152} can slightly outperform FPT on \textbf{ResNet-50}, it has more parameters.
\begin{table*}[t]
\scriptsize
\begin{center}
\renewcommand\arraystretch{1.3}
\setlength{\tabcolsep}{1.8pt}{
\begin{tabular}{cc|c:cc:cc:cc:cc:cc:cc}
Methods & Backbone & \multicolumn{2}{c}{AP} & \multicolumn{2}{c}{AP$_{50}$} & \multicolumn{2}{c}{AP$_{75}$} & \multicolumn{2}{c}{AP$_S$} & \multicolumn{2}{c}{AP$_M$} & \multicolumn{2}{c}{AP$_L$} & Params\\ \hline \hline
BFP+ FPT & ResNet-50 & {38.0} & 36.8 & 57.1 & 55.9 & 38.9 & 38.6 & 20.5 & 18.8 & 38.1 & 35.3 & 55.7 & 54.2 & 88.2 M \\ \hline 
BFP+ FPN~\cite{lin2017feature} & ResNet-101 & 36.2 & 35.7 & 59.1 & 58.0 & 39.0 & 37.8 & 18.2 & 15.5 & 39.0 & 38.1 & 52.4 & 49.2 & 88.0 M \\ 
BFP+ BPA~\cite{liu2018path} & ResNet-101 & 37.3 & 36.3 & 60.4 & 58.7 & 39.9 & 38.3 & 18.9 & 16.3 & 39.7 & 39.0 & 53.0 & 50.5 & 88.4 M  \\ \hline
BFP~\cite{lin2017feature} & ResNet-152 & 35.8 & 34.6 & 55.7 & 53.8 & 37.8 & 35.6 & 15.3 & 14.3 & 35.2 & 33.2 & 51.5 & 45.8 & 89.3 M \\
BFP+ FPN~\cite{lin2017feature} & ResNet-152 & 38.3 & 37.1 & 60.2 & 58.5 & 39.7 & 38.0 & 19.0 & 16.1 & 39.6 & 38.9 & 53.0 & 50.1 & 91.2 M \\
\end{tabular}}
\end{center}
\caption{Result comparisons on different backbones. Experiments are carried out on validation set of MS-COCO 2017~\cite{lin2014microsoft}. ``BFP'' is the bottom-up feature pyramid (BFP)~\cite{lin2017feature}. Results on the left and right of the dashed are of bounding box detection and instance segmentation.}
\label{tab:r3}
\end{table*}
\section{More Qualitative Results}\label{sec:s5}
\begin{figure*}
\centering
\includegraphics[width=1 \textwidth]{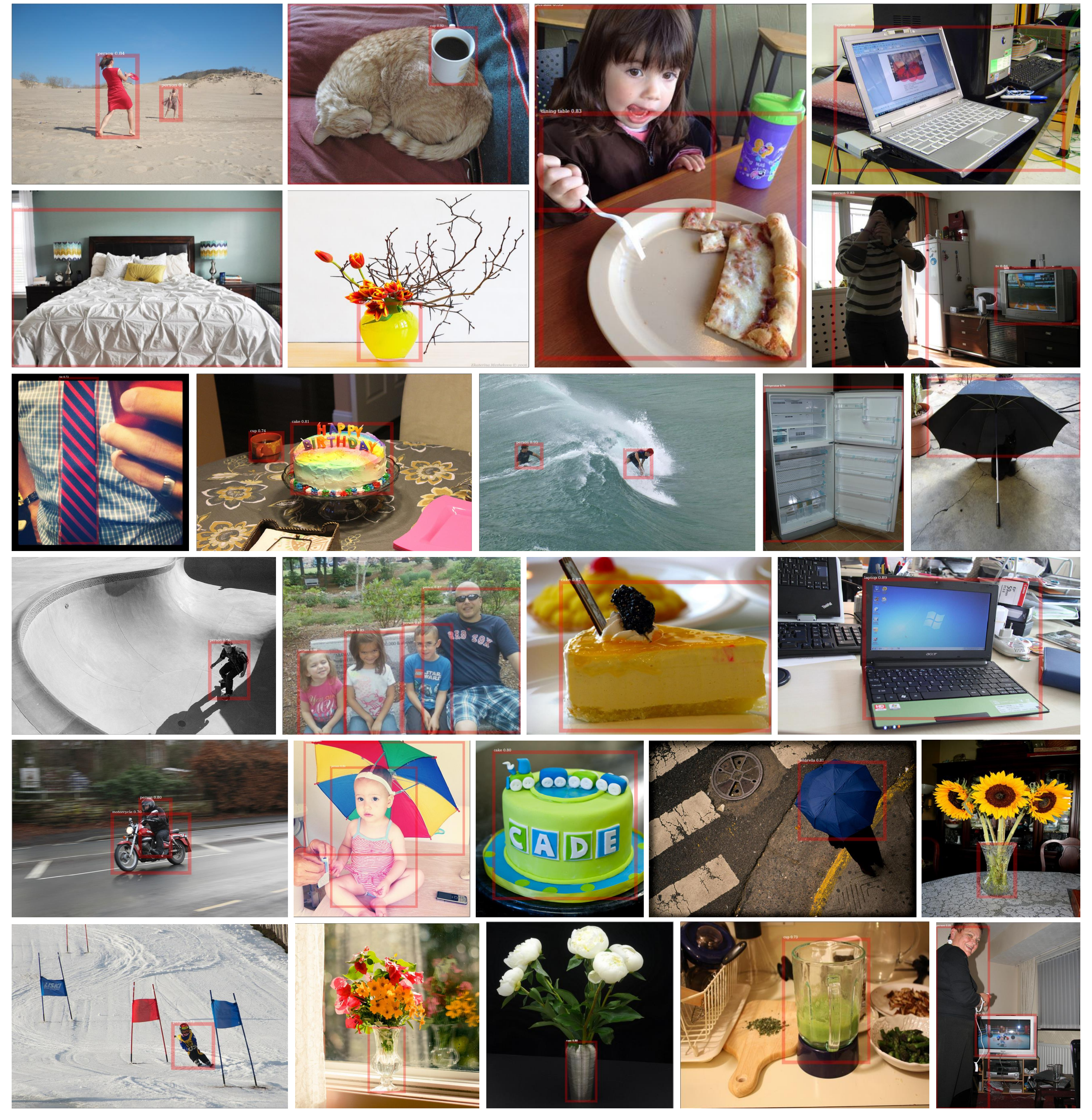}
\caption{\textbf{More object detection results.} Samples are from test set of MS-COCO 2017~\cite{lin2014microsoft}.}
\label{fig:s1}
\end{figure*}

\begin{figure*}
\centering
\includegraphics[width=1 \textwidth]{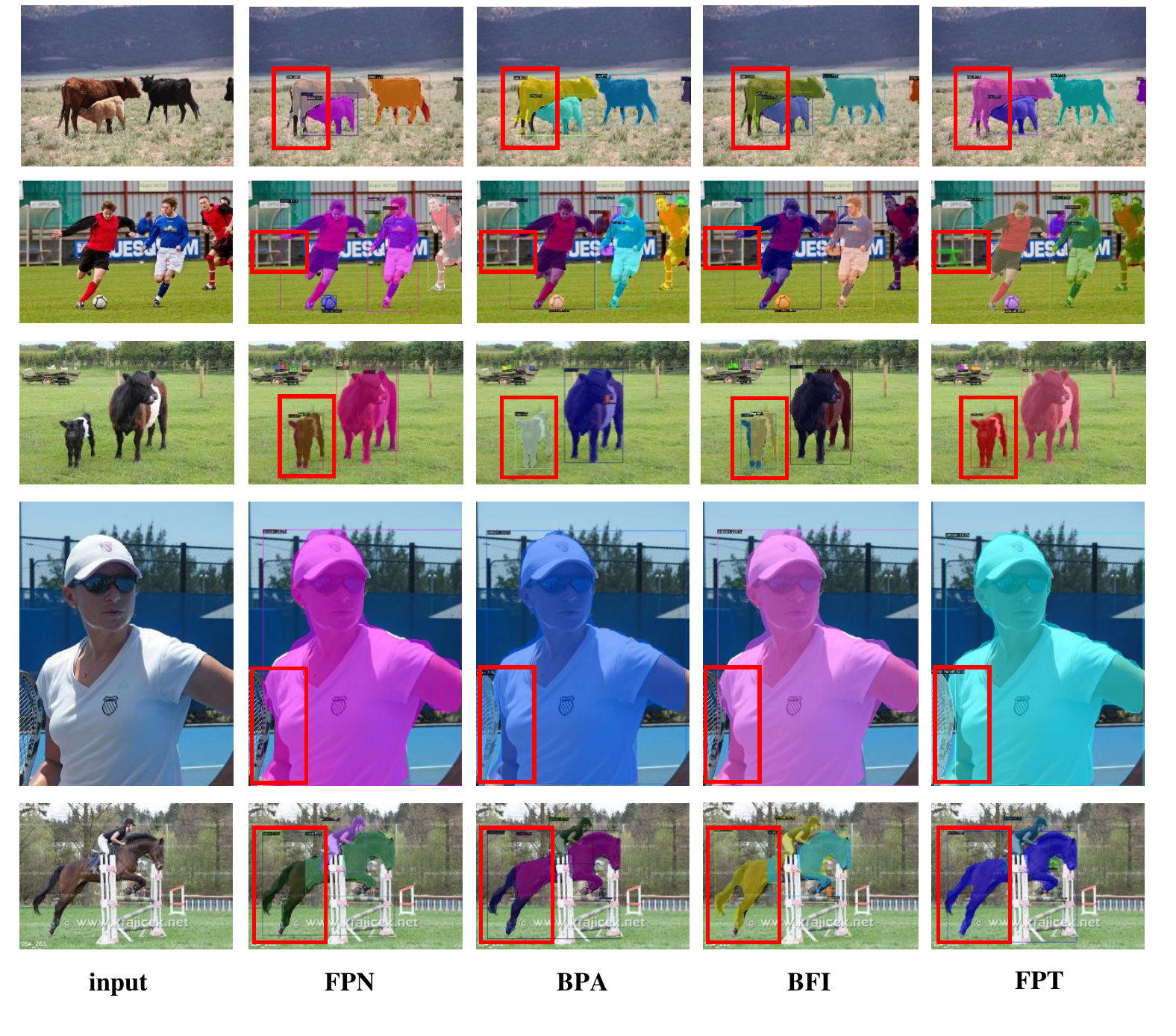}
\vspace{-10mm}
\caption{\textbf{More instance segmentation results.} Samples are from test set of MS-COCO 2017~\cite{lin2014microsoft}. The red rectangle highlights the better predicted areas of FPT.}
\vspace{-8mm}
\label{fig:s2}
\end{figure*}

\begin{figure*}
\centering
\includegraphics[width=.9 \textwidth]{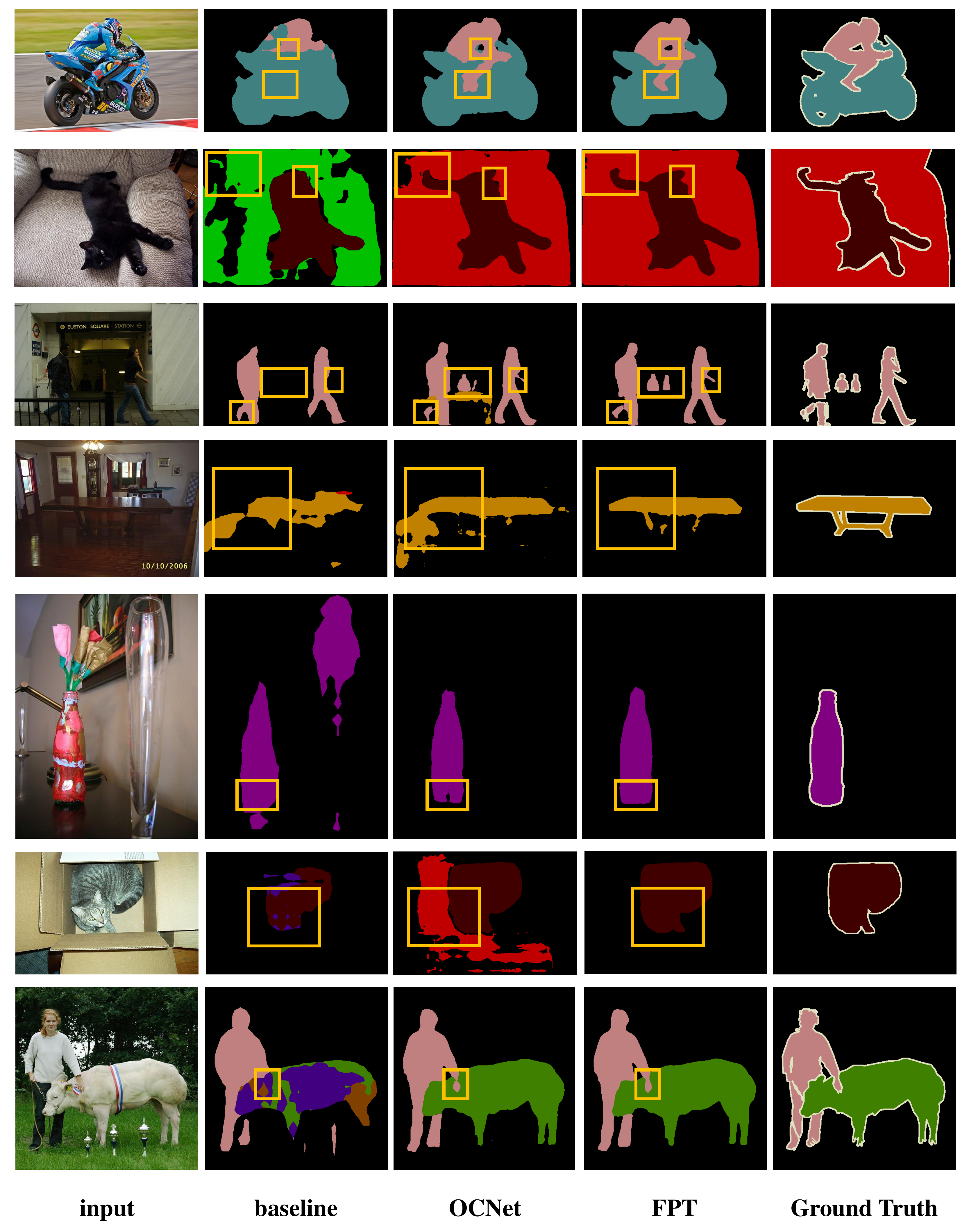}
\caption{\textbf{Semantic segmentation results.} Samples are from val set of PSACAL VOC 2012 ~\cite{everingham2015pascal}. The yellow rectangle highlights the better predicted areas of FPT.}
\label{fig:s3}
\end{figure*}

This section supplements to the visualization results given in Section 4.1.1 and Section 4.2.2 (of the main paper). The results of object detection, instance segmentation and semantic segmentation are visualized respectively in Fig.~\ref{fig:s1}, Fig.~\ref{fig:s2} and Fig.~\ref{fig:s3}. The samples for object detection and instance segmentation are from the test set of MS-COCO 2017~\cite{lin2014microsoft}. As we can see in Fig.~\ref{fig:s1} and Fig.~\ref{fig:s2} that most of our predictions are of high quality, \eg, small objects such as persons and sheep in the distance are correctly detected. The samples for semantic segmentation are from the validation set of PSACAL VOC2012~\cite{everingham2015pascal}. The demonstration in Fig.~\ref{fig:s3} validates that FPT achieves precise segmentation of the thinner objects, \eg, the biker's foot, the cat's tail, the man in the distance and the woman's arm. Moreover, FPT enhances the segmentation quality of larger objects, \eg, the sofa, the bottle, and the dining-table.

\begin{figure*}
\centering
\includegraphics[width=1 \textwidth]{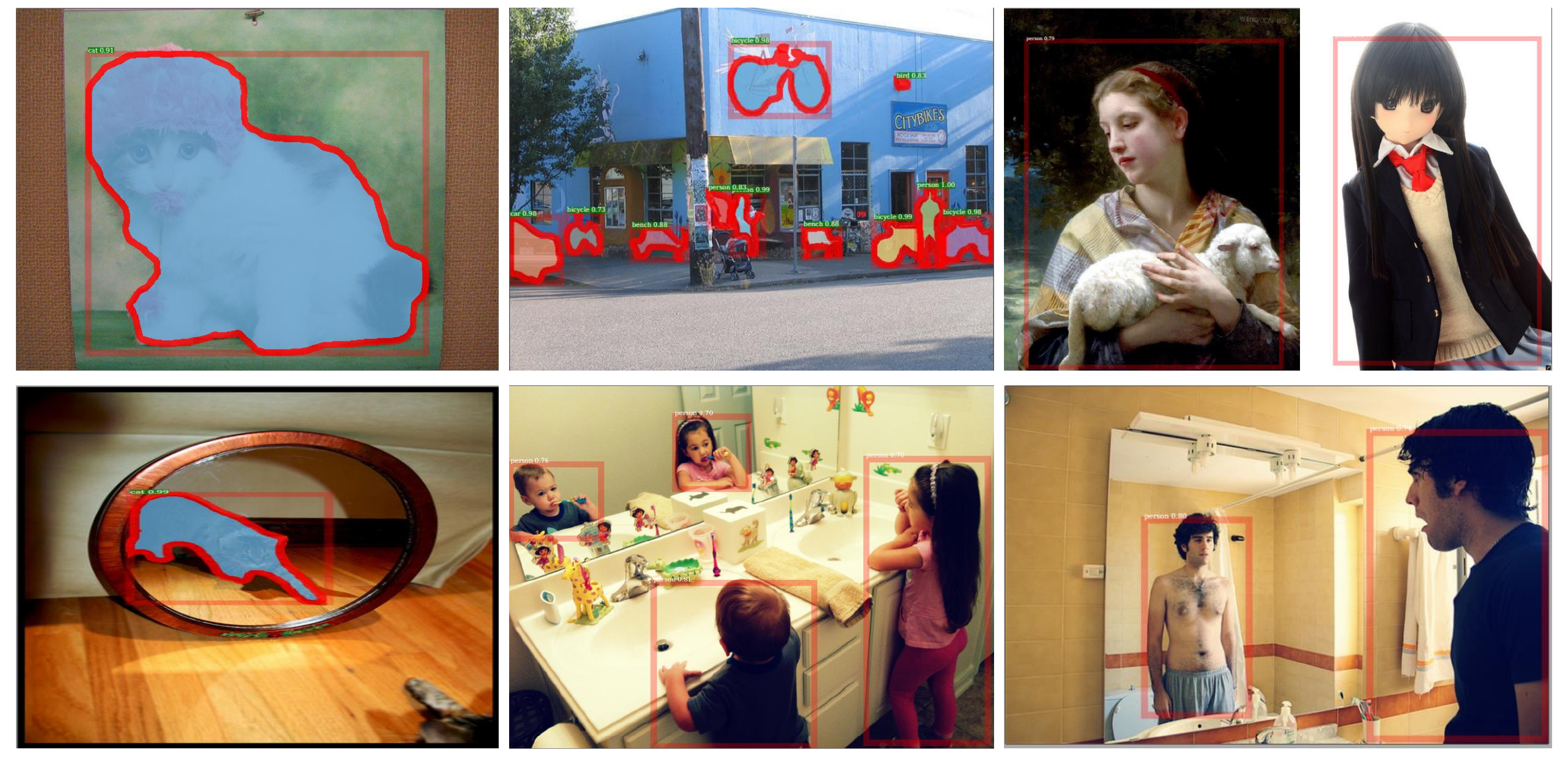}
\vspace{-5mm}
\caption{\textbf{Failure examples of object detection and instance segmentation.} Samples are from test set of MS-COCO 2017~\cite{lin2014microsoft}.}
\vspace{-5mm}
\label{fig:s4}
\end{figure*}

In Fig.~\ref{fig:s4}, we additionally present the failure examples for object detection and instance segmentation.
One possible reason for these failure results is that the background of these objects is not annotated in the ground truth, for example there is no category information for ``mirror'' and ``painting'' in the MS-COCO 2017~\cite{lin2014microsoft} dataset. Hence, objects in these backgrounds can easily be recognized as the real ones, \eg, the cat in the painting, the bike on the wall, and the man in the mirror.

\end{document}